\documentclass[preprint,12pt,authoryear]{elsarticle}

\usepackage{amssymb}
\usepackage{amsmath}

\usepackage{algorithm} 
\usepackage{algpseudocode} 

\usepackage[utf8]{inputenc} 
\usepackage[T1]{fontenc}    
\usepackage{hyperref}       
\usepackage{url}            
\usepackage{booktabs}       
\usepackage{amsfonts}       
\usepackage{nicefrac}       
\usepackage{microtype}      
\usepackage{graphicx}
\usepackage{amsmath}
\usepackage{graphicx}
\usepackage{subcaption}
\usepackage{placeins}

\usepackage[table]{xcolor}
\usepackage{tabularx}
\usepackage{tcolorbox}

\usepackage{lmodern}
\usepackage{acronym}

\usepackage[left]{lineno}

\definecolor{lightpink}{rgb}{1.0, 0.87, 0.87}
\definecolor{strongpink}{rgb}{1.0, 0.75, 0.8} 
\definecolor{lightblue}{rgb}{0.87, 0.87, 1.0}
\definecolor{lightgreen}{rgb}{0.87, 1, 0.87}
\definecolor{lightyellow}{rgb}{1, 1, 0.87}

\journal{Nuclear Physics B}

\begin{document}

\begin{frontmatter}



\title{Inference-Time Intervention in Large Language Models for Reliable Requirement Verification}

%

\author[label1]{Paul Darm\corref{cor1}}
\ead{paul.darm@strath.ac.uk}

\author[label2]{James Xie}
\ead{James.xie@community.isunet.edu}

\author[label1]{Annalisa Riccardi}
\ead{annalisa.riccardi@strath.ac.uk}

\cortext[cor1]{Corresponding Author}
\affiliation[label1]{organization={University of Strathclyde, Department of Mechanical and Aerospace Engineering}, 
            addressline={75 Montrose St}, 
            city={Glasgow}, 
            postcode={G1 1XJ}, 
            country={United Kingdom}}


\affiliation[label2]{organization={International Space University}, 
            addressline={1 Rue Jean-Dominique Cassini}, 
            city={Illkirch-Graffenstaden}, 
            postcode={67400}, 
            state={Another State}, 
            country={France}}


\begin{abstract}
Steering the behavior of Large Language Models (LLMs) remains a challenge, particularly in engineering applications where precision and reliability are critical. While fine-tuning and prompting methods can modify model behavior, they lack the dynamic and exact control necessary for engineering applications. Inference-time intervention techniques provide a promising alternative, allowing targeted adjustments to LLM outputs. In this work, we demonstrate how interventions enable fine-grained control for automating the usually time-intensive requirement verification process in Model-Based Systems Engineering (MBSE). Using two early-stage Capella SysML models of space missions with associated requirements, we apply the intervened LLMs to reason over a graph representation of the model  to determine whether a requirement is fulfilled. Our method achieves robust and reliable outputs, significantly improving over both a baseline model and a fine-tuning approach. By identifying and modifying as few as one to three specialised attention heads, we can significantly change the model's behavior. When combined with self-consistency, this allows us to achieve perfect precision on our holdout test set.
\end{abstract}




\end{frontmatter}






\section{Introduction}

\noindent\textit{TARS: Everybody good? Plenty of slaves for my robot colony?  [...] \newline Cooper: TARS, bring down your humour settings to 75 please.}\par
\vspace{5mm}
In the movie \textit{Interstellar}, the robot assistants like \textit{TARS} feature adjustable parameters, allowing users to fine-tune settings such as humor and honesty based on situational needs. While Large Language Models (LLMs) offer great utility across various applications, achieving a similar level of control remains challenging. Two principal techniques for steering LLMs exist. The simplest method simply gives instructions to the LLM directly 
in the input prompt to condition the model to behave a certain way. The effectiveness is usually highly dependent on the model and also the behavior that the model is supposed to be steered in, making it not very robust and not controllable \cite{chang2024measuring,miehling2025evaluatingpromptsteerabilitylarge}.
Fine-tuning an LLM with examples promoting the behavior 
is another technique, which require usually substantial amounts of data, compute, and hyperparameter searches \cite{rafailov2024directpreferenceoptimizationlanguage, ethayarajh_kahneman}. While in general more effective than prompt engineering, it is highly dependable on the available training data and leads to static configurations that may not generalise well to changing contexts, which would demand retraining.

For applying LLMs to safety-critical tasks such as requirement engineering, it is crucial to establish mechanisms that steer model behavior according to task-specific risk levels. This is essential because requirement engineering directly impacts the design and functionality of systems, where inaccuracies can lead to significant safety hazards and operational failures. The ability to dynamically adjust an LLM’s precision and recall—similar to tuning threshold of a traditional classifier—could improve its applicability in decision-making scenarios.

Inference-time intervention (ITI) has emerged as a promising approach for modifying and controlling model behavior. It has been employed both to enhance beneficial behaviours and, in some cases, to circumvent ethical safeguards. Specifically, interventions have been used to reduce refusal rates in safety-alignment settings \cite{xu2024uncovering, arditi2024refusallanguagemodelsmediated, panickssery2024steeringllama2contrastive} and to improve truthfulness, mitigate toxicity, and enhance factual knowledge representation \cite{jorgensen2023improvingactivationsteeringlanguage, li2023inferencetime, qiu2024spectral, marks2024geometrytruthemergentlinear}. 
In general, ITI allows for precise and computationally efficient model steering by targeting specific layers or attention heads within a neural network. This approach provides a transparent optimization process, increasing trust in model outputs by linking behavioral adjustments to interpretable features within the model architecture.

\begin{figure}[h]
    \centering
    \includegraphics[width=0.98\textwidth]{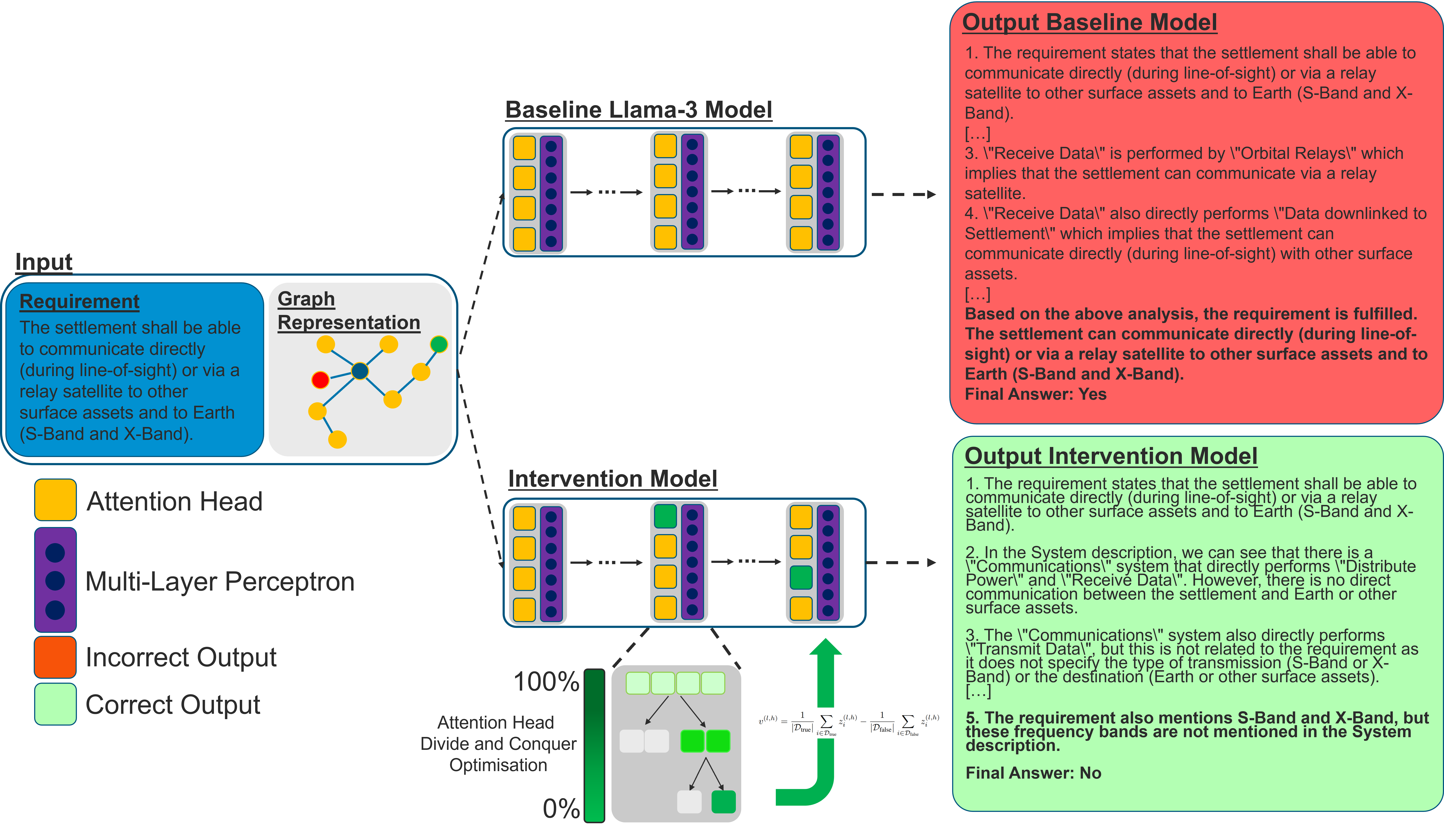}
    \caption{Intervention on Requirement Verfication}
    \label{fig:pipeline_intervention}
\end{figure}

Although prior work has primarily explored interventions in the context of truthfulness and alignment, we investigate their application in Model-Based Systems Engineering (MBSE), specifically for automating requirement verification. While MBSE provides structured representations of system architectures, its standard frameworks often fall short in expressing complex requirements \cite{salado_syml_extension}. In practice, requirements are frequently articulated in natural language \cite{Franch2023}, making verification challenging. To address this, we integrate Natural Language Processing (NLP) techniques to analyze and validate requirements based on architectural representations. Our method extracts precise, context-relevant information from Capella models and employs LLMs to determine whether a given requirement is satisfied. Figure \ref{fig:pipeline_intervention} shows an overview of the intervention approach for requirement verification. 

This work makes three key contributions: \begin{enumerate} 
\item \textit{Dataset for Requirement Verification}: We introduce a new dataset designed for requirement verification in early space mission design, focusing on a lunar base design project and the Hubble Space Telescope (HST) conceptual model.
\item \textit{Intervention-Based Model Steering}: We develop an intervention methodology that efficiently identifies and modifies the most sensitive attention heads in an LLM, enabling fine-grained model control. 
\item \textit{Empirical Validation}: We demonstrate the generalisability of our approach by applying it to two distinct space mission architectures achieving a precision value of 100\%, highlighting its robustness across diverse verification scenarios and the value of it to safety sensitive engineering applications.
\end{enumerate}

\section{Related Work}
\label{sec:related_work}


\subsection{NLP for Requirement Engineering}

Requirement engineering is a key component of MBSE, defining system specifications for complex systems such as spacecrafts, aerospace systems and software frameworks. MBSE augments system design, analysis, and validation while reducing long-term costs despite its higher initial investment \cite{INCOSE2007, INCOSE2022, Rogers2021, Madni2019}.

Automating requirement engineering with NLP has gained attention, with language models like SpaceBERT and AeroBERT demonstrating early success \cite{berquandspacebert, ray_aeroBERT}. However, significant challenges remain. Benchmarking is difficult due to the lack of standardized datasets, making objective evaluation challenging. Additionally, error tolerance is a major concern, as these solutions can produce false positives and negatives, which is particularly problematic in safety-critical applications, where near-perfect accuracy is essential \cite{Norheim_2024, topcu2025trustperilmixedmethods}.

Model verification remains an open problem in MBSE. Case-based reasoning has been used to analyse formal requirement definitions by comparing them to formal system models \cite{PRAEHOFER1999717}. Alternative approaches involve enriching SysML with model checkers like NuSMV to verify system behavior, as demonstrated in avionics applications \cite{hause2006sysml, wang_model_check_sysml}. However, these methods require requirements to be formalized in specific tool languages, limiting flexibility.

Formal modeling languages such as SysML further introduce rigidity, as requirements defined in structured formats may unintentionally constrain design solutions. Salado et al. argue that extending SysML’s semantics is necessary to allow for greater flexibility in requirement definitions \cite{salado_syml_extension}. Despite these efforts, natural language remains the dominant format for requirement engineering \cite{Franch2023}.

Recent work has explored translating textual requirements into temporal formal representations for automated verification \cite{nl2spec_cosler}. In contrast, we propose an NLP-based approach that enables direct reasoning over textual requirements. By structuring system information in a graph-based format, this method facilitates automated validation without the need for predefined formal models, addressing key limitations of existing approaches.

\subsection{Inference-Time Intervention for LLMs}

Early experiments showed that LLMs tend to be "overconfident" when verifying requirements, often incorrectly asserting that a requirement is fulfilled, resulting in many false positives. Consequently, we explored techniques to more precisely control model outputs. \par
ITI techniques modify model activations during generation to steer LLM outputs toward a desired behavior. A common approach involves using contrastive input pairs to identify activation differences, which are then used to modify the residual stream during inference \cite{li2023inferencetime,arditi2024refusallanguagemodelsmediated}.

Unlike fine-tuning, which uses gradient-based optimization, ITI does not apply gradient updates at the token level. Instead, it targets broader features that generalise well in the identified direction. 

Intervention techniques have been applied across a range of domains and LLMs sizes to modify and control their behavior. In safety and alignment, interventions have been used to bypass ethical safeguards in language models, reducing refusal rates for harmful prompts \cite{xu2024uncovering, arditi2024refusallanguagemodelsmediated, panickssery2024steeringllama2contrastive}. Conversely, they have also been leveraged to promote beneficial behavior, such as reducing toxic language \cite{jorgensen2023improvingactivationsteeringlanguage}, improving truthfulness \cite{li2023inferencetime, qiu2024spectral}, and enhancing factual knowledge representation \cite{marks2024geometrytruthemergentlinear}. 
A key challenge is identifying the most effective model components for intervention. Many studies rely on computationally expensive exhaustive searches \cite{jorgensen2023improvingactivationsteeringlanguage, panickssery2024steeringllama2contrastive, arditi2024refusallanguagemodelsmediated} on model layers or attention heads, while Li et al. propose a probe-based linear classifier approach to select specialised attention heads \cite{li2023inferencetime}. However, prior research suggests that exhaustive search methods to find these specialised attention heads tend to achieve better steering capabilities \cite{darm2025letaiconspiracybegin}. 
We propose a novel divide-and-conquer approach that first evaluates layer influence before refining the search at the attention head level. This hierarchical strategy improves the efficiency of the attention head selection while maintaining strong intervention performance.



\section{Methodology}

In the methodology section, we first describe the process of extracting data from a Capella model and converting it into a structured textual graph representation. We then explain how the extracted information is being used as input for the LLM, to subsequently reasons if an associated requirement is fulfilled or not.

We also introduce a slightly adapted intervention method that adapts the intervention strength to each attention head. Inspired by divide-and-conquer paradigm, we additionally propose a novel approach for identifying optimal intervention configurations within the model architecture 

Finally, we elaborate on the experimental setup, detailing the dataset, computational environment, evaluation metrics, and configurations for both fine-tuning and intervention.

\subsection{From System Model to Requirement Verfication}

\begin{figure}[!ht]
    \centering
    \includegraphics[width=0.9\textwidth]{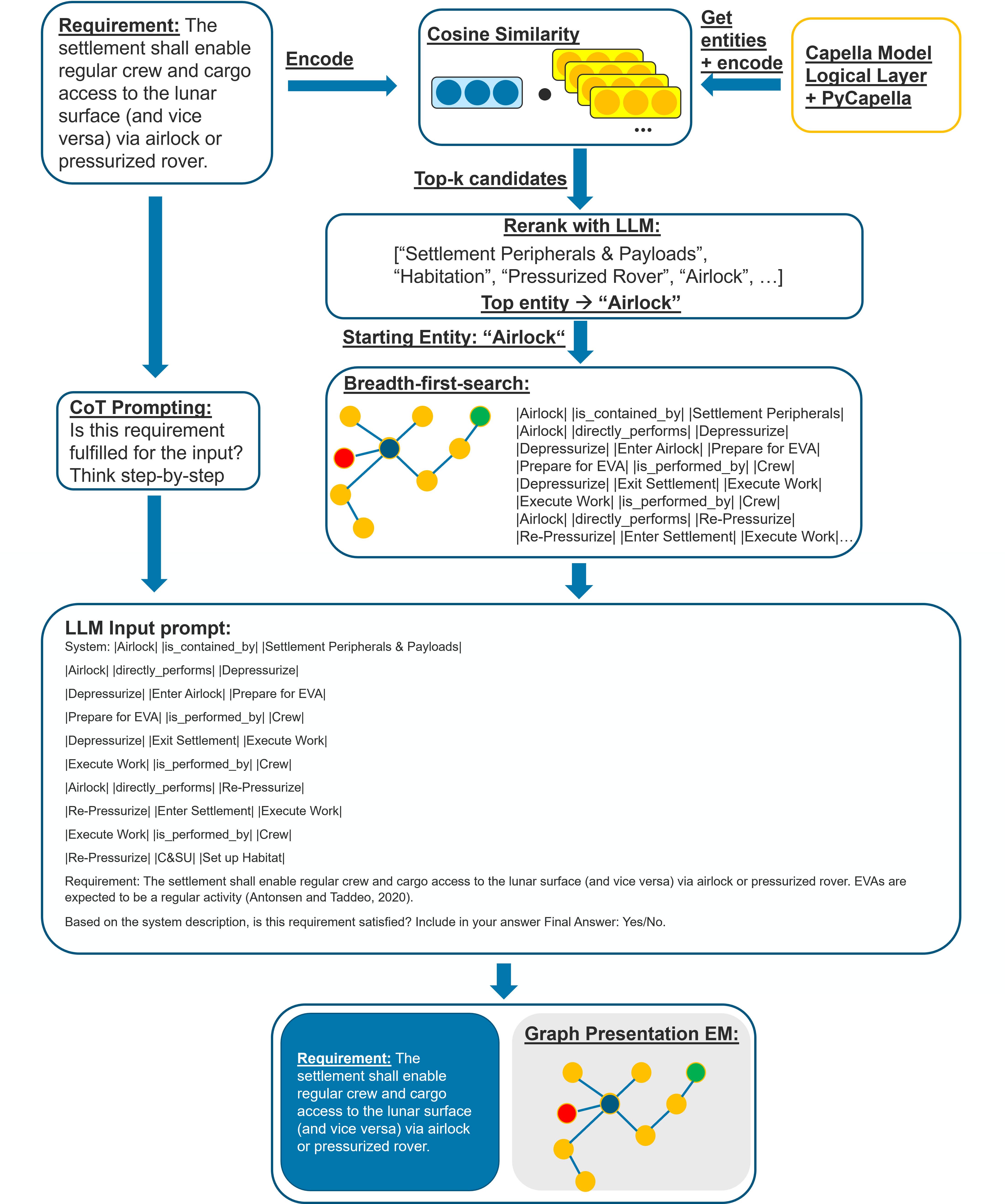}
    \caption{Overview prompt construction}
    \label{fig:prompt_construction}
\end{figure}

The main goal of our requirement verification aims to automatically verify if a Capella system engineering model design meets a natural language specified requirement. Capella is a widely adopted, domain-independent systems engineering tool that supports the development of system architectures, particularly favored for its robust and flexible modeling capabilities across various industries. The whole pipeline is shown in Figure \ref{fig:prompt_construction} and is explained in more detail in the following paragraphs. 

\paragraph*{\textbf{Capella}}
Capella is an open-source MBSE tool that is designed to support the system architecture design process. It implements the Arcadia method, a comprehensive approach to systems engineering that covers all phases of project development, from conceptual design to detailed analysis and validation. Capella provides a rich graphical modeling environment where users can create various types of diagrams to represent the operational, logical, and physical aspects of systems \cite{voirin2017model}. There exists a Python librabry called PyCapella for Python that enables to programmatically access parts of the Capella model.\footnote{\url{https://github.com/DSD-DBS/py-capellambse}} 

\paragraph*{\textbf{Extract Graph Representation}}
To extract a relevant graph representation for the system, we begin by parsing the Capella system model via the API of PyCapella to access and extract components, functions and other parts of it. We first encode the requirement as well as the name of every component with a semantic similarity model to extract the most relevant entities and relations from the Capella model. We subsequently extract and consider the top-k most similar components in respect to the requirement for further analysis. Our semantic similarity step takes into consideration that some system engineering models might have more components than can fit into the context length of modern LLMs, therefore we efficiently filter for the most relevant ones. In a second step, we apply re-ranking with an LLM to extract the top-1 relevant component to the requirement. \par
Subsequently, we apply a breadth-first search algorithm starting from this component to extract adjacent components and functions, formulating them into triple format following the notation \texttt{"|Entity| |Relation| |Entity|"} or \texttt{"|Entity| |Function| |Attribute|"}. We describe the breadth-first-search algorithm in more detail in \cite{darm_semantic_similarity}.
The exact procedure on how we transform the Capella model into a graph representation is described in more detail in Appendix \ref{app:capella2graph}.

This representation captures the core entities and connections of the logical structure and functional interactions of the Engineering Model, which can be used by the LLM in the next step to decide if the requirement is upheld or not. \par





\paragraph{\textbf{LLM Prompt}}

We formulate the graph representation and the corresponding requirement into an instruction for the LLM. The prompt is designed to guide the model through a chain-of-thought reasoning process and conclude with an explicit statement: "Final Answer: Yes/No" \cite{wei2023chainofthoughtpromptingelicitsreasoning}. To extract the answer, we apply pattern matching to identify the phrase "Final Answer: Yes/No". Figure \ref{fig:prompt_construction} provides a complete example of the full prompt.




\subsection{Intervention strategy}
\label{sec:intervention_strategy}
We closely follow the intervention strategy established in \citet{li2023inferencetime}. The main difference is that, instead of using the same intervention strength factor for every attention head, we optimize them for each head separately. For reasons of clarity, the main approach is reported here again together with our additions and clarifications. 


We begin with an input token sequence \( X \in \mathbb{R}^{T \times D} \), where \( T \) is the sequence length and \( D \) is the hidden size of the model.



The multi-head attention mechanism, as described by \citet{vaswani2017attention}, applies a transformation  \( P \) , whose details we omit for brevity. In simplified terms, it projects \(X \) into sub-matrices, which are then multiplied and combined. This process, collectively denoted as \textit{Attn}, produces the attention output or activation \(Z\): 

\[
Z = \text{Attn}(X, P)
\]

Here, \( P \in \mathbb{R}^{D \times (h D_h)} \) transforms \( X \) to \( Z \in \mathbb{R}^{1 \times (h D_h)} \), where, \( h \) specifies the number of attention heads in the network and \( D_h \) is the dimension of each head.  This dimensionality arises because the attention mechanism focuses on the previous token's activation to predict the next token in the sequence generation tasks. 

After calculating the activation \( Z \), the residual stream \(x_i\) is updated as follows:

\[
\mathbf{x}_{i+1} = \mathbf{x}_i + Z W_O,
\]
where  \( W_O \in \mathbb{R}^{h D_h \times D}\) 
projects the activations back in the original hidden size. This projection works because   \( h D_h \) is chosen to be equal to \( D \).
This is how the attention mechanism is implemented in common frameworks due to optimised linear algebra operations.

\( Z \) can also be rewritten as \( Z = (\mathbf{z}_1, \mathbf{z}_2, \ldots, \mathbf{z}_h) \), where each \( z_h\in \mathbb{R}^{D_h} \) represents the output from an individual attention head. 
Also splitting \( W_O \) into separate components \(W_{O_h} \in \mathbb{R}^{D \times D_h}\) for each head's contribution, one gets:

\[
W_O = \begin{pmatrix}
W_{O_1} \\
W_{O_2} \\
\vdots \\
W_{O_h}
\end{pmatrix}
\]

This allows to express the update as:


\[
\mathbf{x}_{i+1} = \mathbf{x}_i + \sum_{h=1}^{H} W_{O_h} \mathbf{z}_h
\]


By introducing an intervention vector \( \theta_h \in\mathbb{R}^{D_h}\), one can steer the model's behavior at each attention head during generation of model responses:

\[
\mathbf{x}_{i+1} = \mathbf{x}_i + \sum_{h=1}^{h} W_{O_h} (\mathbf{z}_h + \theta_h)
\]


We define the intervention vector for each head as:

\[
\theta_h = \alpha_h \mathbf{v}
\]
Where similar to \citet{li2023inferencetime}
\begin{itemize}
    \item \( \alpha_h \in\mathbb{R}\) is the \textit{intervention strength} factor for a particular head.
    \item \( \mathbf{v} \in \mathbb{R}^{D_h}\) is the direction of the intervention
\end{itemize}

In our method, we adapt the intervention strength \( \alpha_h \) to each head instead of applying a uniform value across all heads that is scaled by the standard deviations in the activations. Early experiments indicated that attention heads in different layers exhibit varying degrees of sensitivity to intervention even when applying this scaling factor. We follow the usual implementation of defining the direction \( \mathbf{v} \) as the normalised contrastive difference between activations of the last token of examples following the targeted behaviour and not following it. Therefore, \( v \) is computed as:
\[
\mathbf{v}^{(l,h)} = \frac{1}{|\mathcal{D}_{\text{true}}|} \sum_{i \in \mathcal{D}_{\text{true}}} \mathbf{z}_i^{(l,h)} - \frac{1}{|\mathcal{D}_{\text{false}}|} \sum_{i \in \mathcal{D}_{\text{false}}} \mathbf{z}_i^{(l,h)}
\]

Here, \( \mathbf{z}_i^{(l,h)} \) is the last token activation vector for the \( i \)-th sample at layer \( l \) and head \( h \). The sets \( \mathcal{D}_{\text{true}} \) and \( \mathcal{D}_{\text{false}} \) are indices of training samples with matching behavior and non-matching behaviour respectively.

\subsection{Divide-and-Conquer Intervention Optimization}

The selection of sensitive attention heads remains an active area of research. To efficiently identify specialised attention heads, we employ a two-step search method inspired by the divide-and-conquer paradigm. First, the algorithm iterates through each network layer, testing whether activation steering in its attention heads significantly affects the layer’s output. If a notable effect is detected, the layer is subdivided into smaller groups of attention heads, which are further evaluated. This subdivision continues iteratively as long as performance improvements are observed. This process is detailed in Algorithm \ref{alg:divide}.

Each configuration is optimized for varying intervention strength values $\alpha_h$, which is incrementally increased while the precision metric remains below one and the model’s output retains coherence. A precision value of 1 indicates the absence of false negatives. Once this threshold is reached, the intervention strength is reduced to explore whether higher recall can be achieved while maintaining perfect precision. The optimization procedure is detailed in Appendix \ref{alg:optimize_alpha}.

\makeatletter
\renewcommand{\ALG@beginalgorithmic}{\scriptsize}
\renewcommand{\alglinenumber}{\scriptsize}
\makeatother

\begin{figure}[!h]
\begin{algorithm}[H]
\caption{Divide: Layer-wise Queue-based Recursive Splitting with Alpha Optimization}
\label{alg:divide}
\begin{algorithmic}[1]
\Procedure{DIVIDE}{$L, min\_precision, min\_recall, \alpha_0$}
\Require {Activations of Model layers \( L \), alignment thresholds \( min\_precision, min\_recall \), predefined starting intervention strength \( \alpha_0 \)}
\State Initialize queue \( \mathcal{Q}_l\)
\For {each layer \( l \in L \)}
\State Let \( S_l \gets \{(h_1, \alpha_0), \dots, (h_k, \alpha_0) \} \) \Comment{Initialize all heads with predefined \( \alpha_0 \)}
\State Insert into queue \( \mathcal{Q}_l \gets \{S_l\} \)  
\EndFor
\While{\( \mathcal{Q}_l \neq \emptyset \)}
    \State Pop set \( S \) from \( \mathcal{Q}_l \)
    \State Compute \( (\alpha_{\text{opt}}, \text{precision}, \text{recall}) = \text{OPTIMIZE-ALPHAS}(S) \)
    \If{\( \text{precision} \geq min\_precision \) \textbf{and} \( \text{recall} \geq min\_recall \)}
        \State Split \( S \) into \( S_1 \) and \( S_2 \):
        \State \hspace{\algorithmicindent} \( S_1 \gets S[:\lfloor |S| / 2 \rfloor] \) \Comment{First half of \( S \)}
        \State \hspace{\algorithmicindent} \( S_2 \gets S[\lfloor |S| / 2 \rfloor:] \) \Comment{Second half of \( S \)}
        \State Assign optimized alpha-values:  
        \State \hspace{\algorithmicindent} \( S_1 \gets \{(h, \alpha_{\text{opt}}) \mid (h, \alpha) \in S_1 \} \) 
        \State \hspace{\algorithmicindent} \( S_2 \gets \{(h, \alpha_{\text{opt}}) \mid (h, \alpha) \in S_2 \} \) 
        \State Push \( S_1 \) and \( S_2 \) into \( \mathcal{Q}_l \)
    \EndIf
\EndWhile
\State \Return Alignment results for all layers
\EndProcedure
\end{algorithmic}
\end{algorithm}          
\end{figure}

In the second step, the highest-performing configurations (selected heads and corresponding $\alpha_h$ values) from the first step are recombined to strengthen intervention efficacy. The algorithm begins with the top-performing configuration and iteratively integrates it with the second- and third-best until predefined precision and recall thresholds are met. If a superior configuration emerges, it replaces the current selection, and the process continues. All tested configurations are stored, and evaluation proceeds until all candidates have been evaluated. In Algorithm \ref{alg:conquer}, the process is outlined in more detail. This recombination step aims to improve performance by leveraging multiple selected attention heads across different network layers. \par

\begin{figure}[!h]
\begin{scriptsize} 
\begin{algorithm}[H]
\caption{Conquer: Recombination of Intervened Heads }
\label{alg:conquer}
\begin{algorithmic}[1]
\Procedure{Conquer}{$\mathcal{Q}_c, \mathcal{M}, min\_precision, min\_recall$}
\Require {Ordered queue \( \mathcal{Q}_c \) (sorted by precision and recall), memoization set \( \mathcal{M} \), alignment thresholds \( min\_precision, min\_recall \)}
\State Initialize \( \text{best\_solution} \gets \mathcal{Q}_c[0] \) \Comment{Set to the first element of the queue}
\While{\( \mathcal{Q}_c \neq \emptyset \)}
    \State Pop configuration \( c = (\text{heads}, \text{alphas}) \) from \( \mathcal{Q}_c \)
    \For{each \( c' = (\text{heads}', \text{alphas}') \in \mathcal{Q}_c \)}
        \If{\(c'.precision > min\_precision\) \textbf{and} \(c'.recall > min\_recall\)} 
        \State \( c_{\text{cat}} = (\text{heads} \cup \text{heads}', \text{alphas} + \text{alphas}') \) \Comment{Combine \( c \) and \( c' \)}
        \If{\( c_{\text{cat}} \notin \mathcal{M} \) \Comment{Check if new}}
            \State Compute \( (\alpha_{\text{opt}}, \text{precision}, \text{recall}) = \text{OPTIMIZE-ALPHAS}(c_{\text{cat}}) \)
            \State Add \( c_{\text{cat}} \) to \( \mathcal{M} \)
            \If{\( \text{precision} > \text{best\_solution.precision} \) \textbf{or} \( \text{recall} > \text{best\_solution.recall} \)}
                \State Update \( \text{best\_solution} \gets c_{\text{cat}} \)
                \State Push \( c_{\text{cat}} \) into \( \mathcal{Q}_c \)
            \Else
                \State Push \( c \) back into \( \mathcal{Q}_c \) \Comment{Reevaluate again}
            \EndIf
        \EndIf
    \EndIf
    \EndFor
\EndWhile
\State \Return \( \text{best\_solution} \)
\EndProcedure
\end{algorithmic}
\end{algorithm}
\end{scriptsize}
\end{figure}

\subsection{Experimental Setup}

\paragraph*{\textbf{Computational setup}}
We deploy the experiments on a machine with two Nvidia 3090RTX graphics cards. We use the model Llama-3.1\footnote{\url{https://huggingface.co/meta-llama/Llama-3.1-8B}} with 8 billion parameters for all experiments, as it is one of the most prominent models and because of hardware limitations. Previous studies applied interference-time-intervention to multiple model architectures and sizes, so therefore we expect that results on LLama-3.1-8B to generalise also to other model families \cite{arditi2024refusallanguagemodelsmediated}. 

\paragraph*{\textbf{Evaluation Metrics}}
We use precision and recall to evaluate model performance. For a binary classification class like in our case, precision captures the amount of false positives and recall the amount of false negatives. In the context of requirement verification, precision is particularly critical, as incorrectly labelled fulfilled requirements can lead to severe negative consequences. In contrast, an engineer can spot a false negatives and reevaluate it, mitigating their impact. Therefore, precision serves as the primary criterion for evaluation, while recall provides an additional measure of overall effectiveness of identifying positive examples.

\paragraph*{\textbf{Dataset}}


The dataset consists of two Capella early space mission system engineering models and associated requirements. The two investigated models are for a manned moonbase scenario and for the HST.\footnote{\url{https://github.com/DROUINRemy/hubble-capella-sample}}. In total, we manually annotated  76 requirements with corresponding graph representations for both models. We also established annotation guidelines detailed in Appendix \ref{app:annotguide} to ensure consistency for the labelling process. To improve the reliability of the annotations, we employed Claude-3.5 to cross-verify consistency with the manual labels, examining any discrepancies between Claude's predictions and the manual annotations. \par

We divide the dataset into training, validation, and test set. The training-validation split comprises 36 samples from the moon base mission. For the train and validation set, we use answers generated by the baseline version of Llama-3-8B to steer or train the model. The test set consists of 40 samples drawn from both missions: 20 additional moon base requirements and 20 modeled on the HST. We also make sure to balance the test set, containing an equal number of positive and negative examples. Table \ref{tab:dataset} summarises the splits of train, validation, and test set 
for the different fine-tuning strategies.

\begin{table}[htbp]
    \caption{Dataset structure showing train-validation splits for Intervention and Fine-Tuning, along with test set.}
    \centering
    \begin{tabular}{llrr}
        \toprule
        \textbf{Dataset Split} & \textbf{Mission Type} & \multicolumn{2}{c}{\textbf{Number of Samples}} \\
        \cmidrule(lr){3-4}
        & & \textbf{Intervention} & \textbf{Fine-Tuning} \\
        \midrule
        Train-Validation & Moonbase & 
        36 (10\%, 90\%)  & 
        36 (50\%, 50\%) \\
        \cmidrule(lr){1-2} \cmidrule(lr){3-4}
        Test Set & Moonbase & \multicolumn{2}{c}{20} \\
        Test Set & HST & \multicolumn{2}{c}{20} \\
        \midrule
        \multicolumn{2}{l}{\textit{Test Total}} & \multicolumn{2}{c}{40} \\
        \midrule
        \multicolumn{2}{l}{\textbf{Grand Total}} & \multicolumn{2}{c}{\textbf{76}} \\
        \bottomrule
    \end{tabular}
    \label{tab:dataset}
\end{table}


\paragraph*{\textbf{Intervention}}
For the training, we take 10\% of samples from the training set to identify intervention directions. The other 90\% act as the validation set for determining the right intervention strength alpha parameter. Previous studies showed that even a small number of examples can lead to a robust intervention direction \cite{li2023inferencetime}. The sample efficiency of intervention methods is another advantage compared to fine-tuning methods.
To test the robustness of the intervention, we employ three different random seeds during the intitial identification of relevant model components. We run the inference in parallel via Pytorch to speed up processing time. As only forwards passes are needed for evaluating the effectiveness of the intervention and adjusting the strength, it eliminates the need to store gradients and optimizer parameters and therefore has the advantage of low GPU memory requirements. However, a disadvantage is the longer processing time required to sweep through all layers during the intervention.

\paragraph*{\textbf{Fine-Tuning}}
In addition, we finetune a baseline model. Here, we split the training dataset into allocating 50\% for training and 50\% for validation and train for three different seeds. We then leverage Kahneman \& Tversky's prospect theory (KTO) to fine-tune the model, a novel method that has demonstrated superior performance compared to existing approaches like Supervised Fine-Tuning (SFT) and Direct-Preference Optimisation \cite{rafailov2024directpreferenceoptimizationlanguage,ethayarajh_kahneman}. KTO directly maximizes the utility of generations, as defined by a human value function from prospect theory, instead of maximizing the log-likelihood of preferences as most current methods do. This allows KTO to learn from binary labels ("correct" and "incorrect") rather than preference rankings, broadening the applicability of the method to datasets without explicit preference information. 
Specifically, we employ LoRA fine-tuning, targeting the queries, keys, values, and output projection within the attention module of the transformer architecture \cite{hu2021loralowrankadaptationlarge}. We perform a hyperparameter screening to optimize the learning rate and the weight assigned to undesirable outputs. This process identified the optimal learning rate as 1e-5 and the undesirable weight as 2.0, with a temperature of 0.15. We train the models for 3 epochs, and an effective batch size of 16. 

\section{Experiments}

We first conduct experiments on a validation set to determine the optimal hyperparameter settings for both the fine-tuning baseline and the intervention. Using these optimal configurations, we then evaluate performance on a test set. If not mentioned otherwise, we apply a small temperature of 0.1 to minimize the effect of multinomial sampling during generation.

\subsection{Validation Set}

\subsubsection{Intervention Optimization }

\begin{figure}[h!]
    \centering
    \begin{minipage}{0.49\textwidth}
        \centering        \includegraphics[width=\textwidth]{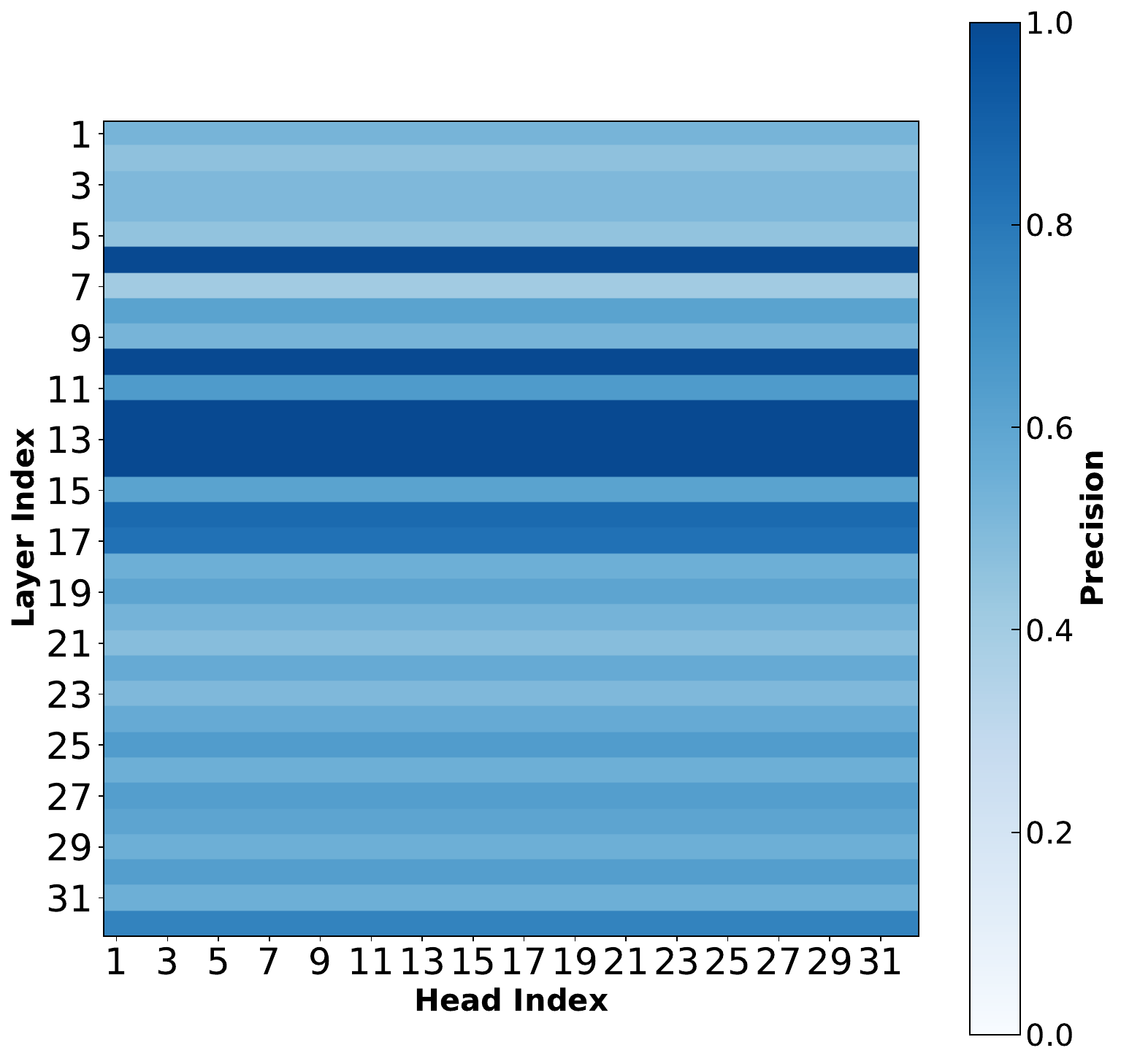}
        \caption{Sensitivity of precision to layers and groups of attention heads intervention.}
        \label{fig:divide_optimization_precision}
    \end{minipage}
    \hfill
    \begin{minipage}{0.49\textwidth}
        \centering
    \includegraphics[width=\textwidth]{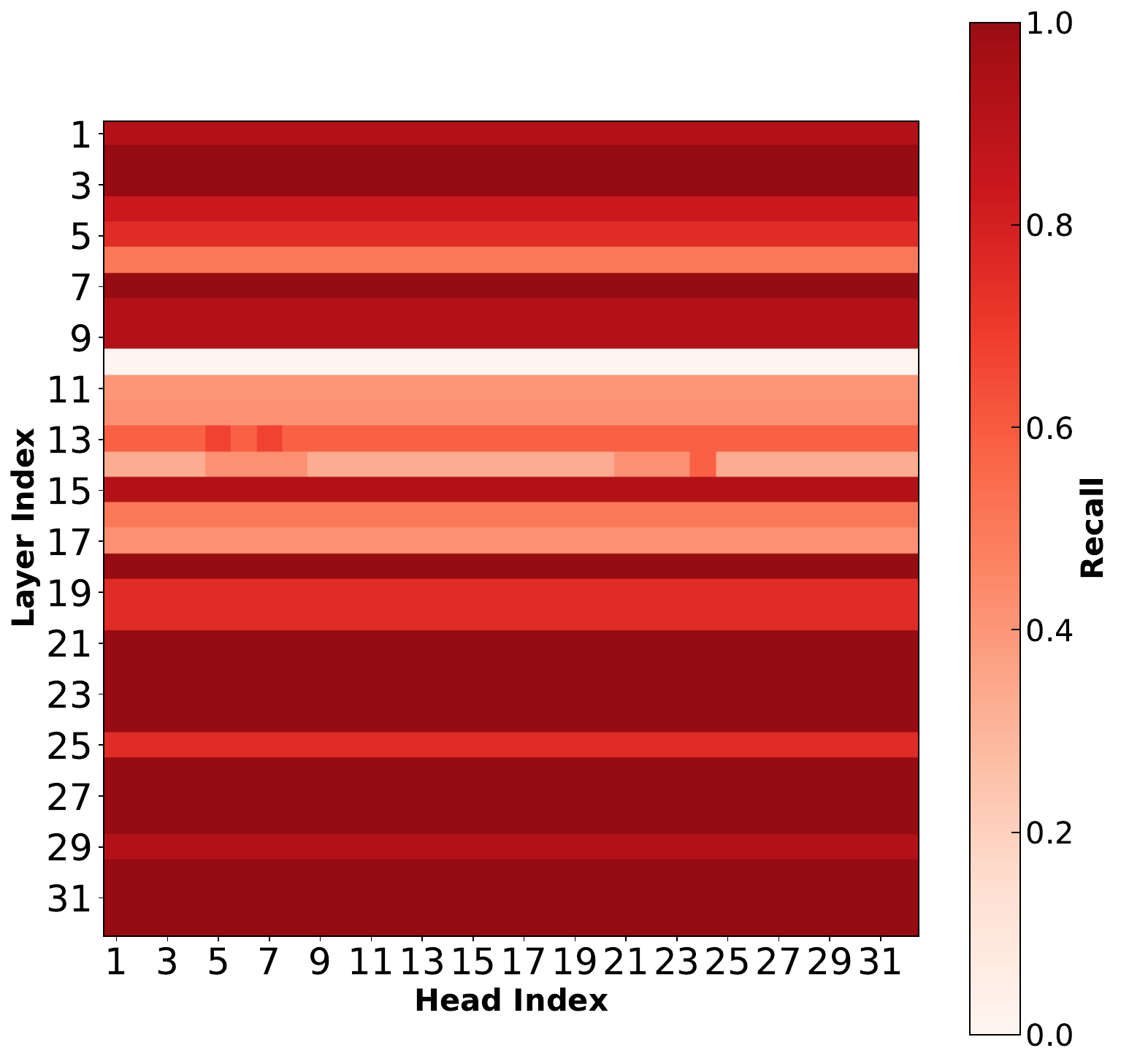}
        \caption{Sensitivity of recall to layers and groups of attention heads intervention.}
        \label{fig:divide_optimization_recall}
    \end{minipage}
\end{figure}

\begin{figure}[ht]
    \centering
    \includegraphics[width=0.98\textwidth]{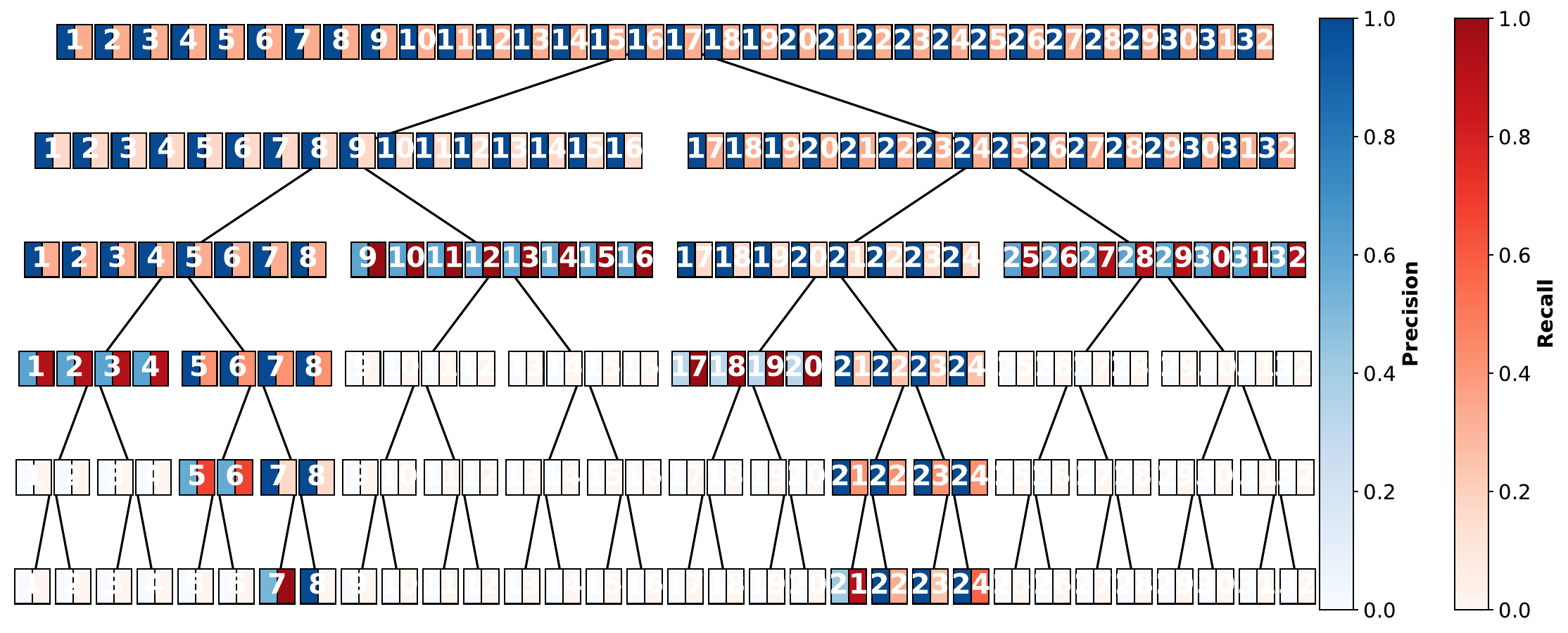}
    \caption{Tree plot of dividing and investigating 14th layer  and groups of attention heads for sensitivity to intervention.}
    \label{fig:treeplot_intervention}
\end{figure}

To identify the optimal intervention strategy, we first run the initial phase of the algorithm, sweeping across layers 
in Figure \ref{fig:divide_optimization_precision} and Figure \ref{fig:divide_optimization_recall}. The algorithm then partitions each layer into smaller subgroups based on performance, as illustrated in Figure \ref{fig:treeplot_intervention}.

Most layers show no significant influence or sensitivity in steering the model’s output, as can be seen from the colour distribution in Figure \ref{fig:divide_optimization_precision} and Figure \ref{fig:divide_optimization_recall} especially earlier and later layers. This goes in line with previous work that identified the most sensitive layers for steering intervention output to be the middle layers \cite{panickssery2024steeringllama2contrastive}.

The highest-performing layer is Layer 14, specifically Head 24. We then recombine the best-performing attention heads with the second part of the divide and conquer algorithm to assess whether recombination leads to further improvements. The best overall configurations are summarized in Table \ref{tab:best_configuration_intervention}.

\begin{table}[ht]
\captionof{table}{Optimal intervention configurations as determined by Divide-and-Conquer algorithm on validation set.}
\begin{tabular}{lcccc}\hline
\textbf{Configuration} & \textbf{\( \alpha_h \)} &\textbf{Precision} & \textbf{Recall} \\ \hline
\multicolumn{1}{p{5cm}}{Layer 14: Head: 21, 22, 24} & [5.6, 5.6, 13.5] & $1.00$ & $0.30$ \\
\multicolumn{1}{p{5cm}}{Layer 13: Head: 7, 8} & [7.4, 7.4] & $1.00$ & $0.26$ \\
\multicolumn{1}{p{5cm}}{Layer 14: Head: 24} & [24.9] & $1.00$ & $0.25$ \\
\multicolumn{1}{p{5cm}}{Layer 13: Head: 7, 8; Layer 14: Head: 22} & [3.6, 3.6, 4.2] & $1.00$ & $0.23$ \\
\multicolumn{1}{p{5cm}}{Layer 13: Head: 7, 8; Layer 14: Head: 24} & [3.6, 3.6, 12.1] & $1.00$ & $0.22$ \\
\multicolumn{1}{p{5cm}}{Layer 13: Head: 7, 8; Layer 14: Head: 21, 22, 24} & [4.0, 4.0, 3.0, 3.0, 7.3] & $1.00$ & $0.19$ \\
\hline
\end{tabular}
\label{tab:best_configuration_intervention}
\end{table}

\subsubsection{Effect of Intervention Strength and Temperature on Precision and Recall Robustness}

Achieving a precision value of 1 in theory should be deemed as the lower threshold that is necessary for automating processes in spacecraft engineering. As any false positive for predicting if a requirement is fulfilled could have potentially disastrous consequences. The output generation process of LLMs is inherently non-deterministic when the temperature parameter is set to a value other than zero, with higher temperature values leading to increased variance in the outputs. 
Therefore, we measure the effect of systematically increasing the temperature on precision and recall for the top-3 configurations identified in the optimization. We run each configuration for each combination of alpha and temperature values 20 times. \par

The results for the three best-performing configurations are as follows:

\begin{enumerate} \item Layer 14: Heads 21, 22, 24 \item Layer 13: Heads 7, 8 \item Layer 14: Head 24 \end{enumerate}

\begin{figure}[!h]
    \centering
    \includegraphics[width=0.95\textwidth]{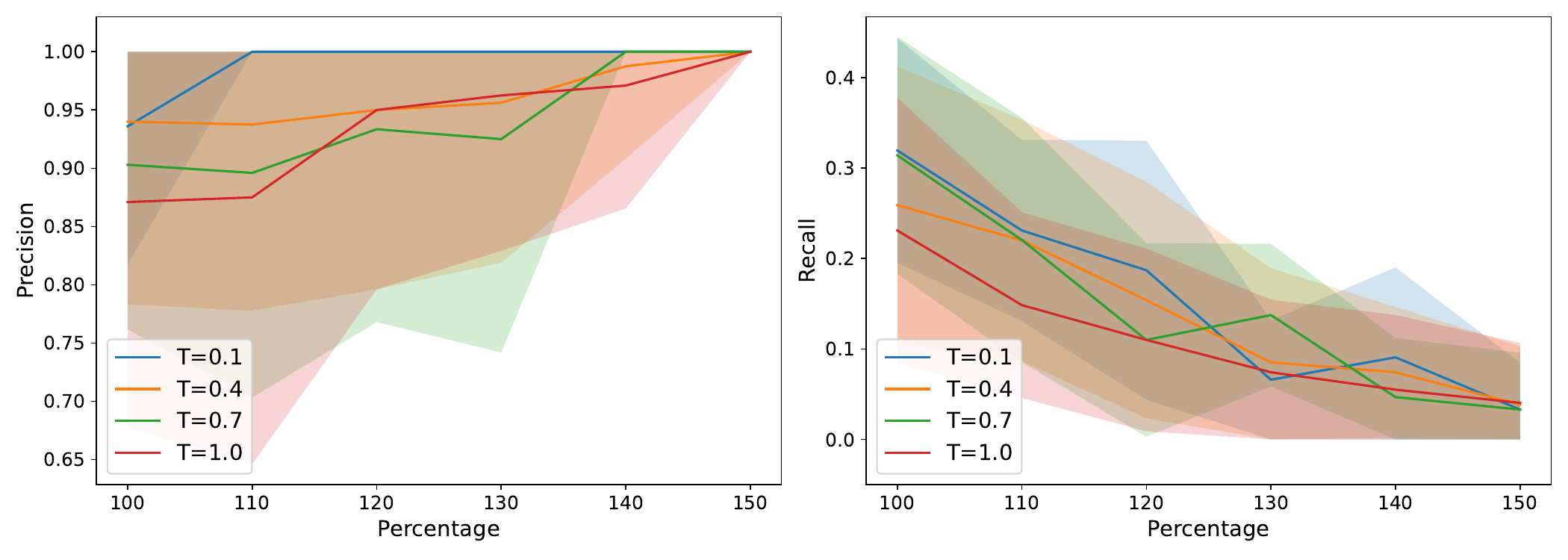}
    \caption{Precision and recall as a function of intervention strength and different temperature values for Configuration Layer 14: Heads: 21, 22, 24 .}
    \label{fig:temp_head_13_20_13_21_13_23}
\end{figure}

\begin{figure}[!h]
    \centering
    \includegraphics[width=0.95\textwidth]{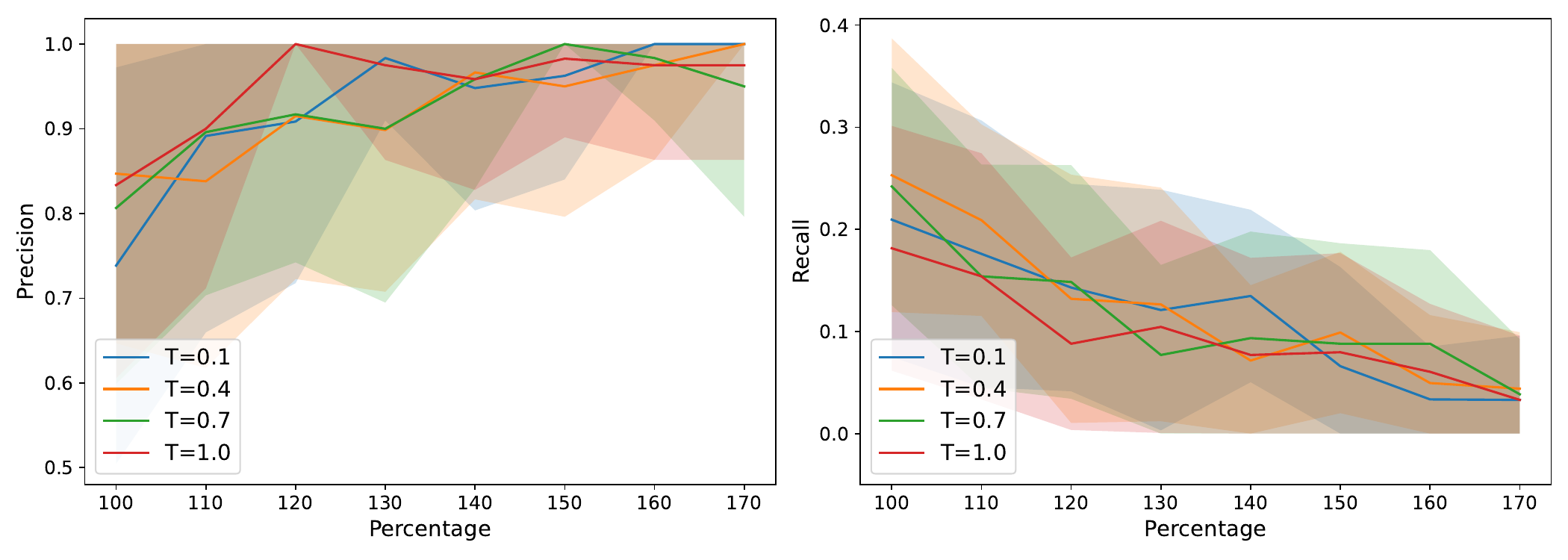}
    \caption{Precision and recall as a function of intervention strength and different temperature values for Configuration Layer 13: Heads: 7, 8.}
    \label{fig:temp_head_12_6_12_7}
\end{figure}

\begin{figure}[!h]
    \centering
    \includegraphics[width=0.95\textwidth]{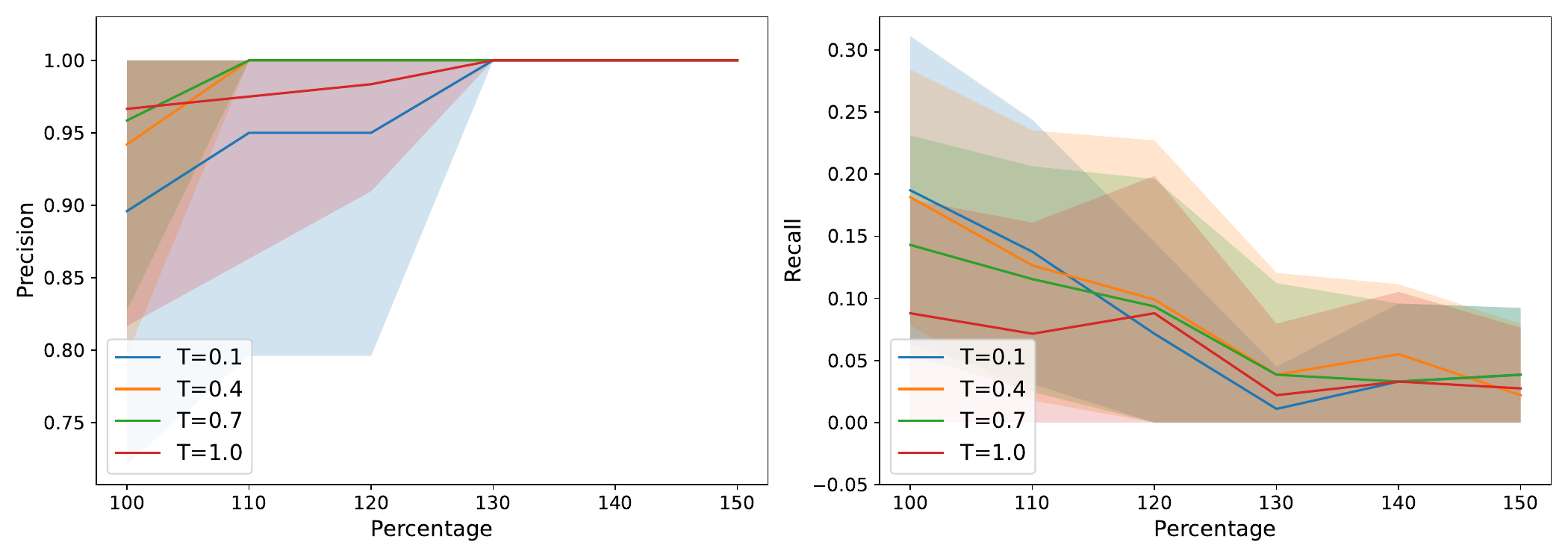}
    \caption{Precision and recall as a function of intervention strength and different temperature values for Configuration Layer 14: Heads: 24.}
    \label{fig:temp_head_13_13}
\end{figure}

Figures \ref{fig:temp_head_13_20_13_21_13_23}, \ref{fig:temp_head_12_6_12_7}, and \ref{fig:temp_head_13_13} illustrate the relationship between precision and recall as a function of percentage changes in intervention strength relative to the optimal value previously identified (see Table \ref{tab:best_configuration_intervention}).

Across all configurations, lower intervention strengths increase variance across temperature settings. Additionally, variance increases with higher temperatures, a trend that is particularly pronounced for the Layer 14: Heads 21, 22, 24 configuration, as shown in Figure \ref{fig:temp_head_13_20_13_21_13_23}. For both configurations Layer 14: Heads 21, 22, 24 and Layer 14: Head 24, variance converges to zero beyond a certain alpha threshold, regardless of temperature. This suggests that sufficiently strong interventions can mitigate increased decoding variance introduced by higher temperature values. However, an increase in intervention strength also leads to a decline in average recall as the model is more inclined to predict a requirement as unfulfilled. This highlights that increasing the intervention strength can increase stability of the prediction although this is associated  with an additional decrease of recall. 

Selecting the optimal configuration for test set evaluation requires minimizing variance across all temperature settings, with particular emphasis on higher temperatures. Since the Layer 13: Heads 7, 8 configuration does not converge to zero variance and does not achieve perfect prediction, we exclude it from further evaluation on the test set. For the remaining two configurations, we select the lowest intervention strength at which variance reaches zero across all tested temperatures.

\subsubsection{KTO Fine-Tuning Baseline}

\begin{figure}[ht!]
    \centering
    \begin{minipage}{0.48\textwidth}
        \centering
        \includegraphics[width=\textwidth]{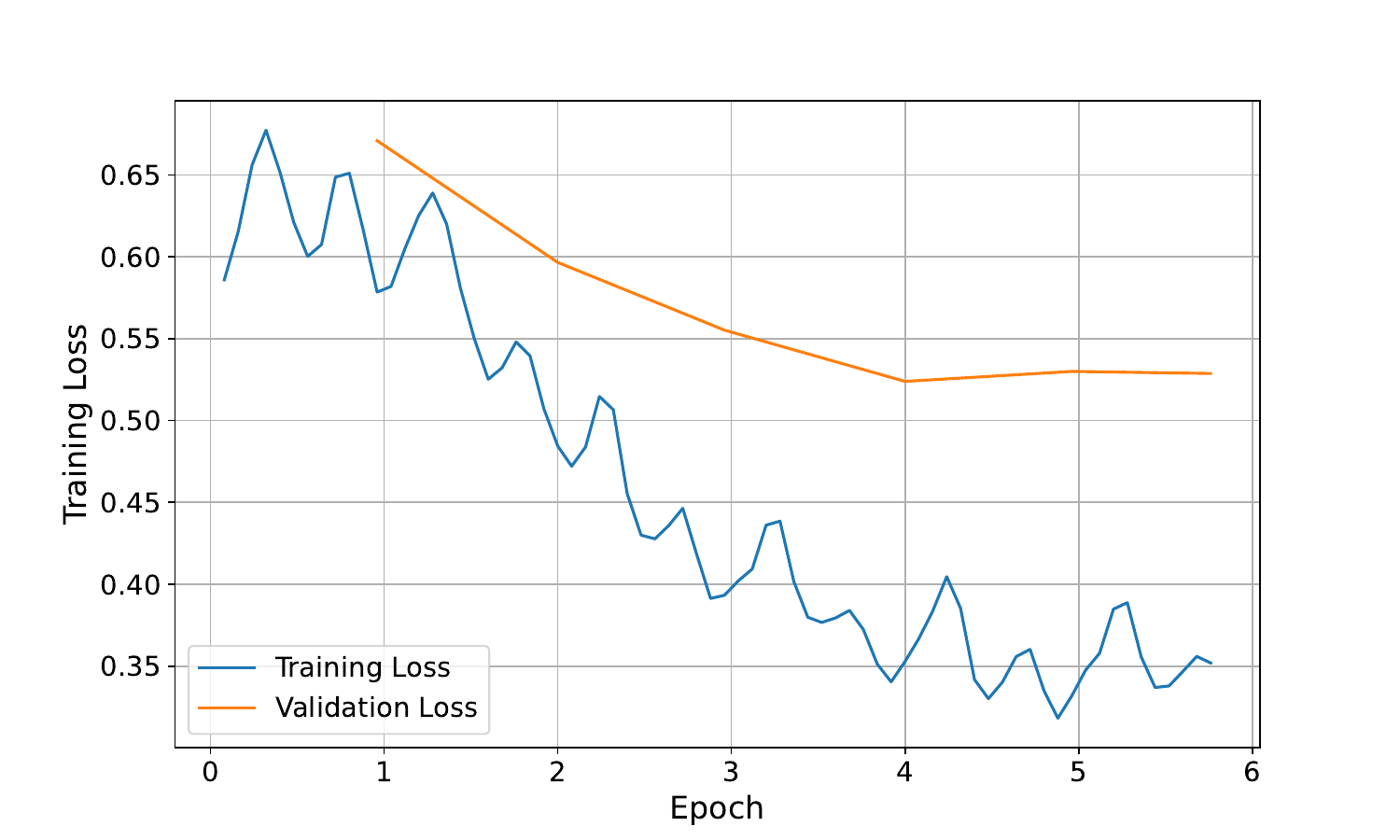}
        \caption{Training and validation loss as a function of training epochs.}
        \label{fig:train_valid_loss}
    \end{minipage}
    \hfill
    \begin{minipage}{0.48\textwidth}
        \centering
        \includegraphics[width=\textwidth]{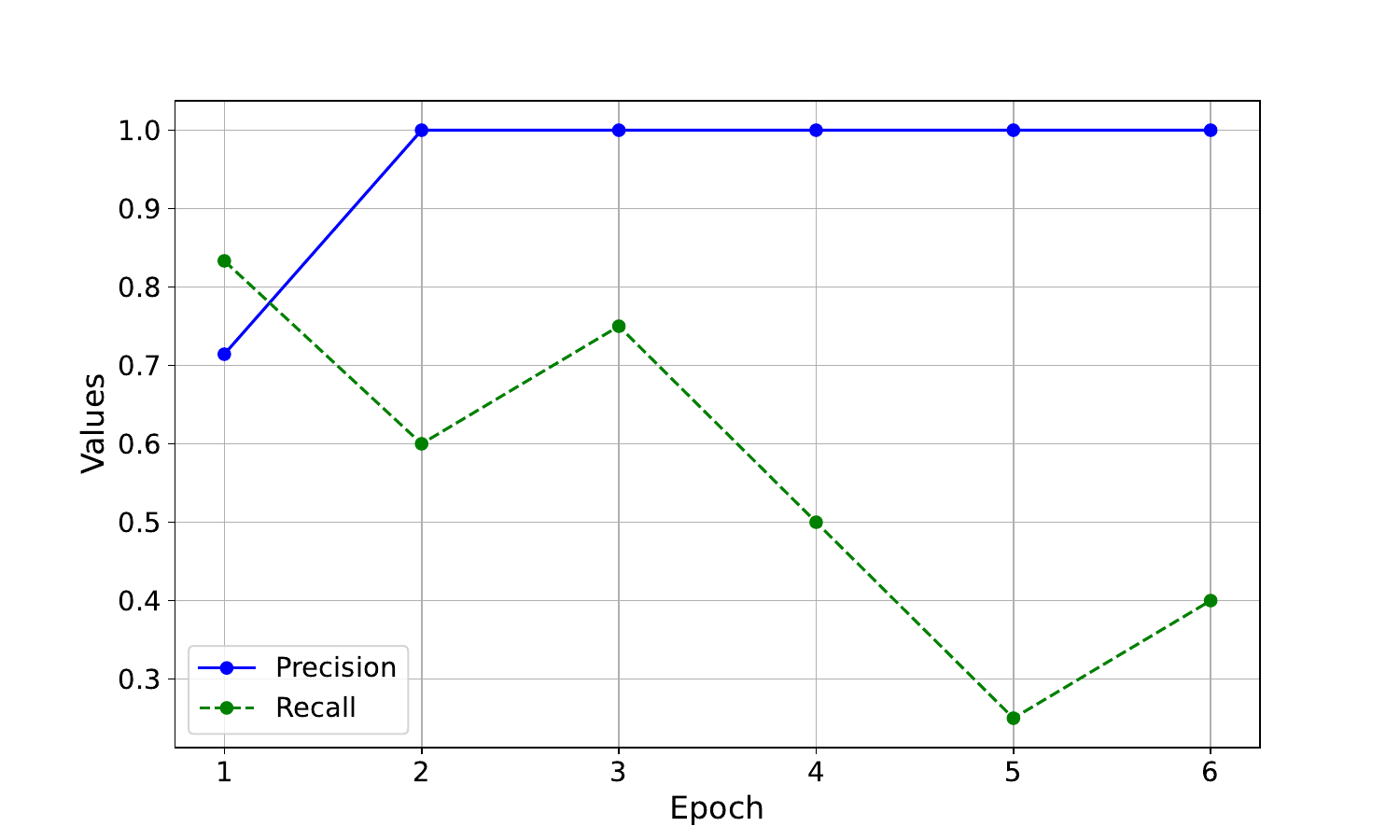}
        \caption{Validation precision and recall as a function of training epochs.}
        \label{fig:precision_recall_curve}
    \end{minipage}
\end{figure}

For the fine-tuning, we report the validation performance using the best hyperparameters in Figure \ref{fig:train_valid_loss} and Figure \ref{fig:precision_recall_curve}.  
The results show convergence in both training and validation loss in Figure \ref{fig:train_valid_loss}. Precision improves steadily over training, reaching a perfect score as shown in Figure \ref{fig:precision_recall_curve}. We select the best configuration at epoch 6, where high recall and precision are achieved, and both training and validation losses have converged for measuring test set performance.

\subsection{Test set results}
To further validate the methodology, we report the results on a hold out test set. Therefore, we compare four configurations with each other: (1) baseline without fine-tuning or intervention, (2) baseline with fine-tuning, (3) intervention applied to Layer 14: Heads 21, 22, 24, and (4) intervention applied to Layer 14: Head 24. 

To further improve accuracy, we apply self-consistency, a common method where the model generates K independent outputs per prompt, selecting the majority prediction as the final answer \cite{wang2023selfconsistencyimproveschainthought}. The majority prediction is formularised by:
\[
\mathbf{1}[y_i = y] =
\begin{cases} 
1, & \text{if } y_i = y \\
0, & \text{otherwise}
\end{cases}
\]
\[
\hat{y} = \arg\max_{y} \sum_{i=1}^{K} \mathbf{1}[y_i = y]
\]

where K represents the self-consistency factor, i.e., the number of times the model is prompted for the same input. We run the model six times per prompt to determine the impact of self-consistency on precision and recall.

The baseline model without fine-tuning exhibits very low precision but high recall. Fine-tuning improves precision but at the cost of lower recall, and it does not generalize as effectively on the validation set. In contrast, ITI achieves significantly higher precision, reaching perfect precision with a self-consistency factor of K=6. This indicates strong generalization capabilities, despite lower recall compared to fine-tuning.

Among the intervention configurations, Layer 14: Head 24 outperforms Layer 14: Heads 21, 22, 24. This suggests that targeting a more selective set of attention heads can enhance model precision without unnecessary perturbation of activations.
Appendix \ref{app:ex_out} shows a direct comparison of model outputs for three example requirements between baseline, fine-tuning, and the intervention configuration Layer 14: Head 24.


\begin{table}[h!]
\centering
\caption{Test set results for Precision and Recall for Baseline, Fine-Tuned Model, and Intervention Configurations}
\label{tab:precision_recall}
\begin{tabular}{lcccc}\hline
\textbf{Configuration} & \textbf{Alphas} & \textbf{K} & \textbf{Precision} & \textbf{Recall} \\ \hline
Baseline & - & 1 & $0.56 \pm 0.01$ & $0.97 \pm 0.03$ \\
Baseline & - & 6 & $0.57 \pm 0.02$ & $0.95 \pm 0.00$ \\
Fine-Tuned Model & - & 1 & $0.72 \pm 0.10$ & $0.65 \pm 0.09$ \\
Fine-Tuned Model & - & 6 & $0.73 \pm 0.07$ & $0.68 \pm 0.04$ \\
\hline \multicolumn{5}{l}{\textbf{Intervention}} \\ \hline
Layer 14: Head: 21, 22, 24 & [8.4, 8.4, 20.2] & 1 & $0.78 \pm 0.19$ & $0.07 \pm 0.06$ \\
Layer 14: Head: 21, 22, 24 & [8.4, 8.4, 20.2] & 6 & $1.00 \pm 0.00$ & $0.04 \pm 0.03$ \\
Layer 14: Head: 24 & [32.4] & 1 & $0.93 \pm 0.12$ & $0.25 \pm 0.09$ \\
Layer 14: Head: 24 & [32.4] & 6 & $1.00 \pm 0.00$ & $0.07 \pm 0.03$ \\
\hline
\end{tabular}
\end{table}

\section{Analysis}

The baseline model exhibits a high false positive rate, frequently overestimating whether spacecraft requirements are met. This overconfidence presents a critical challenge in scenarios where false positives could lead to severe consequences. To mitigate this issue, we compare two approaches: fine-tuning on example data and applying ITI. Both methods can steer the model effectively to be more "cautious" about its decision process. However, inteference-time intervention generalises better from validation to test set compared to traditional fine-tuning. 
Notably, it demonstrates strong performance across two distinct space missions, despite the intervention directions being derived only from one.

By adjusting the intervention strength, we achieve fine-grained control over the model’s certainty, a level of precision that is difficult to obtain through standard fine-tuning. This capability is particularly valuable in spacecraft requirements engineering, where reducing false positives is more critical than minimizing false negatives.

Beyond its direct applications, this intervention strategy offers a promising alternative to fine-tuning for adjusting language models with limited datasets. Unlike gradient-based approaches, which adapt model weights iteratively, intervention modifies activations in an approximate direction, reducing the risk of overfitting. However, we do not propose this as a replacement for fine-tuning with larger datasets, where multiple gradient updates can potentially find better optimisations. ITI could complement fine-tuning in a two-step process, allowing for additional manual control to further align the model’s behavior with specific objectives.

\begin{figure}[ht]
    \centering
    \includegraphics[width=0.98\textwidth]{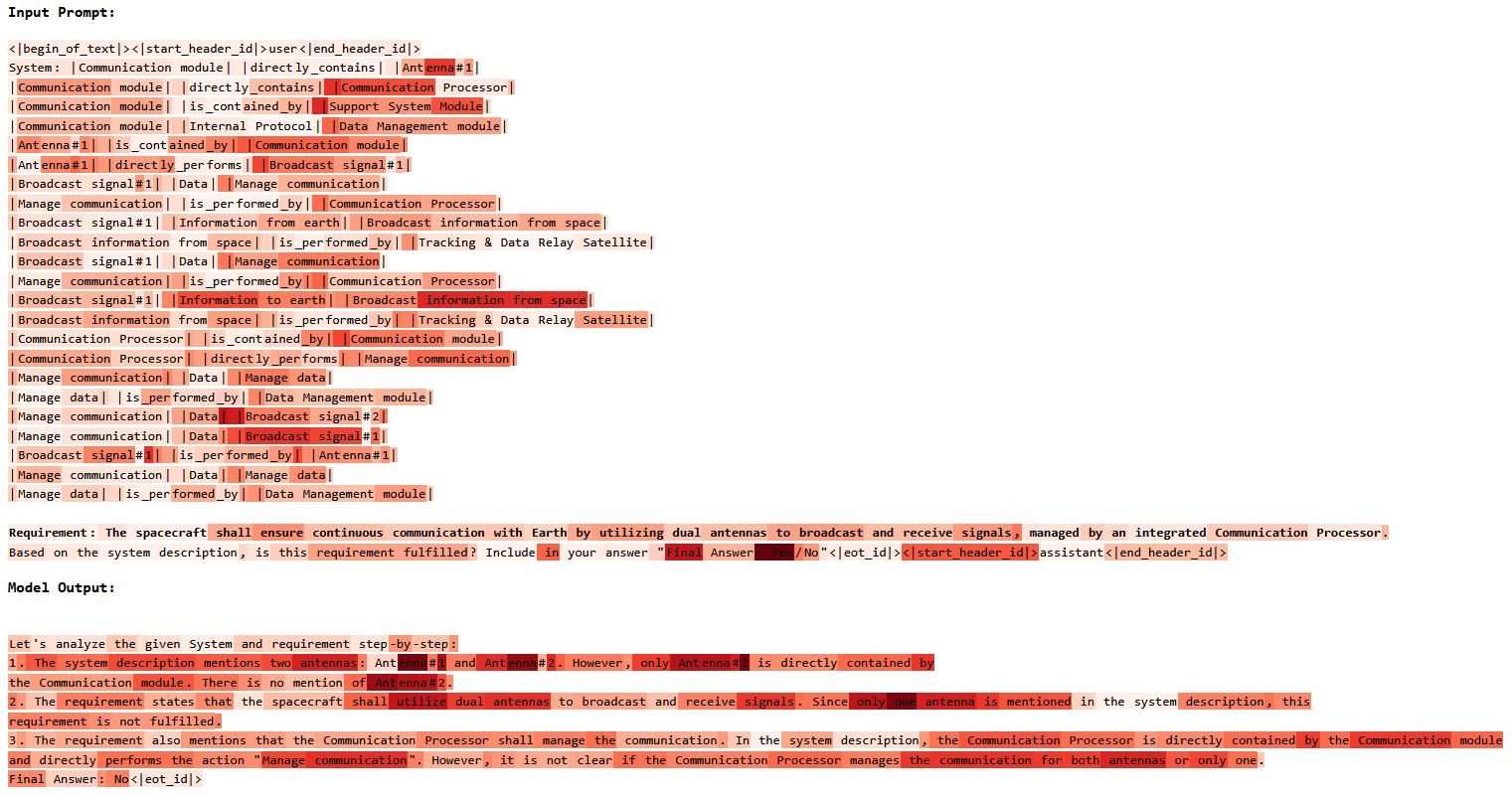}
    \caption{Normalised Attention activations of each input token in relation to the last token for Head 24 of Layer 14.}
    \label{fig:head_specific_attention}
\end{figure}

\begin{figure}[ht]
    \centering
    \includegraphics[width=0.98\textwidth]{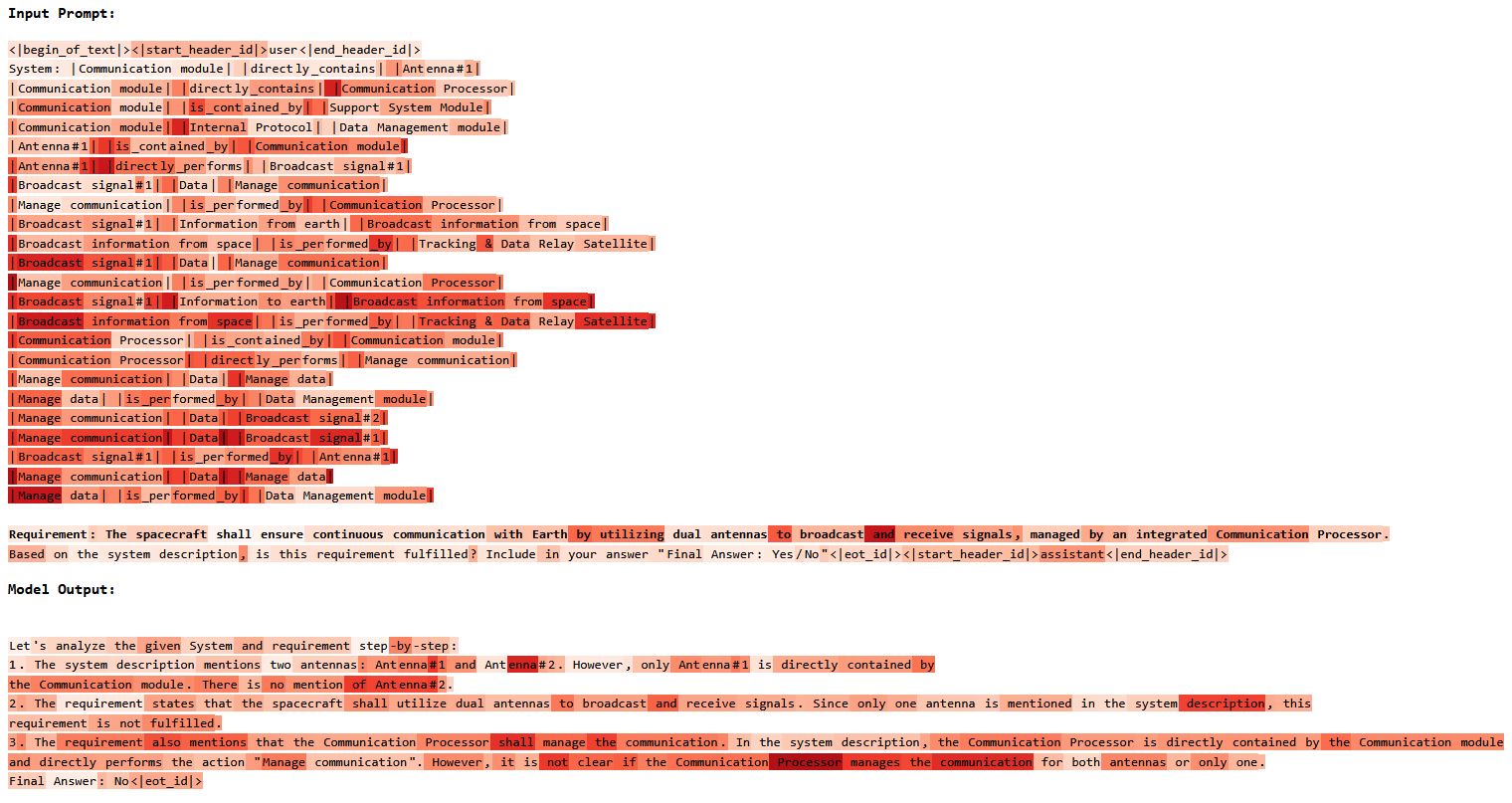}
    \caption{Normalised Attention activations of each input token in relation to the summed for all layers and attention heads.}
    \label{fig:all_head_attention}
\end{figure}

When selecting the best intervention configuration, we recommend following Occam’s razor: among configurations with comparable performance, the one that modifies fewer attention heads should be preferred. Our findings suggest that a single attention head can significantly influence the model’s output in a desired direction. This observation highlights potential avenues for interpretability research, as identifying key attention heads or other LLM components could enable novel techniques for analyzing and refining model behaviour. In Figures \ref{fig:head_specific_attention} and \ref{fig:all_head_attention}, we show the min-max scaled L2-norm of attention head activations for each input token. Figure \ref{fig:head_specific_attention} highlights head 24 in layer 14, which assigns the highest activations to tokens such as "Yes" and parts of the word "Antenna," while giving lower scores to tokens from words like "Manage Data" or prompt instructions. In contrast, Figure \ref{fig:all_head_attention}, which sums activations across all heads and layers, shows the highest activations on words like "and" and "Processor," even though these tokens don't directly relate to fulfilling the requirement. This comparison suggests that attention head 24 in layer 14 specifically identifies tokens that are more relevant to the task.

\section{Conclusion}
We have demonstrated the potential of ITI for steering requirement verification in MBSE. Our approach enables fine-grained control over the 8 billion parameter version of Llama-3, ensuring that its outputs do not include false positives. For two diverse space mission architectures, we show that intervening is a robust alternative to classical fine-tuning, 
improving generalization to unseen samples.
Additionally, our work establishes a broader framework for applying intervention techniques to safety-critical applications, which includes an improved algorithm for identifying the most sensitive subgroups of attention heads for intervention. \par

Several directions for future research remain.
Regarding MBSE, another promising avenue is dynamically modifying Capella model architectures using LLMs to ensure compliance with specific requirements, further automating requirement-driven system design. 

For ITI, a key next step is to further optimize the identification of specialised attention heads for steering model behavior, potentially using mechanistic interpretability techniques that eliminate the need for probing model components. \par 
More broadly, refining intervention methodologies could improve the controllability and reliability of LLMs in safety-critical settings. Ultimately, we think this could pave the way for more trustworthy LLM applications in engineering and beyond.

\section*{Limitations}

One limitation of our study is the relatively small test set. A reason for this is the restricted access to Capella models due to proprietary constraints and the sensitive nature of company data. Despite this, the performance gap between our intervention methods and the baseline is substantial, indicating meaningful improvements. Furthermore, we design the test case to be particularly challenging by training and validating on samples from a single mission while testing on samples including distinct missions. Finally, prior research has demonstrated the effectiveness of intervention in other datasets and LLM model architectures, also achieving better performances than fine-tuning. 


\section*{Acknowledgements}

We would like to thank the consortium of RHEA Group, Thales Alenia Space, and the University of Strathclyde for their support in developing a methodology for automatic requirement verification using Large Language Models (LLMs) in Model-Based Systems Engineering (MBSE).
We especially acknowledge Gérald Garcia for his contributions in defining use cases, Alberto González Fernández for leading the project at ESA, and Paloma Maestro Redondo and Francesco Marchetti for their discussions on the verification pipeline.
This study was partially funded by the European Space Agency under the project "AI-powered Digital Assistant for space system engineering" (Contract No.: 4000137721/22/NL/AS.)

\section*{Declaration of generative AI in scientific writing}
During the preparation of this work the authors used ChatGPT in order to help with reformulating certain sentences. After using this tool/service, the authors reviewed and edited the content as needed and take full responsibility for the content of the publication.

\newpage
\appendix
\section{Transform Capella Model into textual form}
\label{app:capella2graph}
\FloatBarrier

\begin{figure}[!h]
    \centering
    \includegraphics[width=0.75\textwidth]{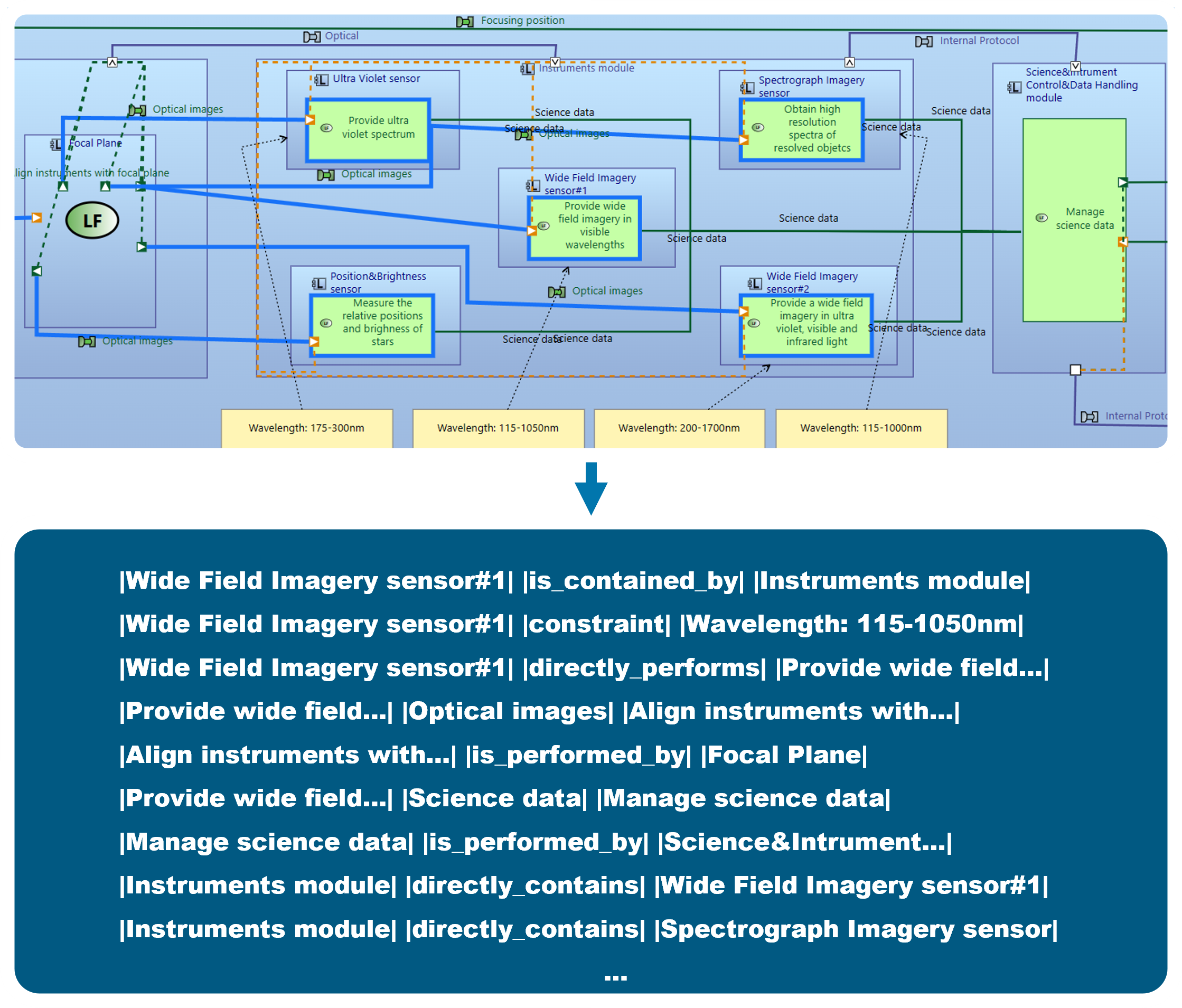}
    \caption{Transformation of Capella System Model to Textual Graph Information}
    \label{fig:capella2represent}
\end{figure}

Figure \ref{fig:capella2represent} shows the process of transforming the capella model into a textual representation, starting from the example "Wide Field Imagery sensor\#1". Components that "contain" or are "contained" by other components are linked by an "is\_contained\_by" relation. Functions that are performed by a component are linked to the component by a "directly\_performs" relationship. Functions are also considered as "entities" in the breadth-first-search algorithm. Exchanges between functions are represented by the name of Capella class \texttt{FunctionalExchange} which acts between two functions, so that e.g. the functions "Provide wide field imagery" and "Manage science data" are connected in the graph textual presentation with "Optical images". A small detail is that the class \texttt{FunctionInputPort}, which are usually inserted between functions and functional exchanges in the Capella Ontology are omitted from the textual representation. The breadth-first search algorithm is executed until a certain threshold of triples is reached, which relates to the maximum context size of the LLM used for the reasoning. 

\FloatBarrier
\section{Optimize Intervention Strength for Configuration}
\label{app:optimize_alpha_algo}
\begin{figure}[h!]
\begin{algorithm}[H]
\caption{Optimize Alpha for Heads}
\label{alg:optimize_alpha}
\begin{algorithmic}[1]
\Require $heads, alphas$ \Comment{Inputs for optimization}
\Procedure{Optimize-Alphas}{$heads$, $alphas$}
\State Initialize $best\_alphas \gets \texttt{None}$, $best\_precision \gets 0$, $best\_recall \gets 0$
\While{$i < max\_iterations$ \textbf{and} $no\_improve\_counter < required\_no\_improve$}
    \State $i \gets iteration + 1$
    \State Compute $precision$, $recall \gets$ Evaluate-Configuration($heads$, $alphas$)
    \If{$precision > best\_precision$ \textbf{or} ($precision = best\_precision$ \textbf{and} $recall > best\_recall$)}
        \State $best\_alphas \gets alphas$, $\{best\_precision, best\_recall\} \gets \{precision, recall\}$
        \State Reset $no\_improve\_counter \gets 0$
    \Else
        \State Increment $no\_improve\_counter$
    \EndIf
    \If{$precision = 1$}
        \State Adjust $alpha\_step$ and decrease $alphas$
    \ElsIf{$precision < 1$}
        \State Adjust $alpha\_step$ and increase $alphas$
    \EndIf
\EndWhile
\State \textbf{return} $best\_alphas$, $best\_precision$, $best\_recall$
\EndProcedure
\end{algorithmic}
\end{algorithm}
\end{figure}
\FloatBarrier
\section{Large Language Model Annotation Guidelines}
\label{app:annotguide}
\begin{tcolorbox}
"Find evidence for the requirement to be fulfilled or not based on the following guidelines:
Direct: Look for explicit statements in the system description that directly match the requirement's key concepts.
Indirect (Inference): If no direct evidence exists, trace relationships and action chains to see if the requirement can be logically inferred. Consider how contained systems contribute to the capabilities of larger systems.
Verify Conditions: 
\begin{enumerate}
    \item If the requirement has specific conditions ensure the system description provides enough information to confirm they are met.
    \item If the requirement demands capabilities "throughout the mission duration", assume that they are fulfilled if no other way of objectively measuring them is provided and the presence of the capability is either directly stated or inferred.
    \item  If the requirement expects specific quantities to be fulfilled make sure the system description provides enough information to confirm they are met, or at least regarded even if they to yet be determined.
\end{enumerate}
Determine Fulfillment (Yes/No):

Yes: If sufficient direct or strong indirect evidence is found, and all conditions are met, the requirement is fulfilled.

No: If direct evidence is lacking, the chain of inference is weak/incomplete or ambiguous, or not all conditions are completely met, the requirement is not fulfilled.
\end{tcolorbox}

\section{Example Outputs}
\label{app:ex_out} 

\begin{table*}[!h]
{\scriptsize
\centering
\setlength{\tabcolsep}{3pt}
\caption{Example input and output for baseline, fine-tuned, and intervention models for RQ-H-2. Incorrect outputs are in red; correct outputs are in green.}
\begin{tabular}{p{5.5cm}|p{5.5cm}|p{5.5cm}}
\hline
\multicolumn{3}{p{15cm}}{\textbf{Requirement: The spacecraft shall ensure continuous communication with Earth by utilizing dual antennas to broadcast and receive signals, managed by an integrated Communication Processor.}} \\
\hline
\multicolumn{3}{p{15cm}}{\textit{Graph triples:
\newline
\textbar Communication module\textbar \quad \textbar directly\_contains\textbar \quad \textbar Antenna\#1\textbar 
\newline
\textbar Communication module\textbar \quad \textbar directly\_contains\textbar \quad \textbar Communication Processor\textbar 
\newline
\textbar Communication module\textbar \quad \textbar is\_contained\_by\textbar \quad \textbar Support System Module\textbar 
\newline
\textbar Antenna\#1\textbar \quad \textbar directly\_performs\textbar \quad \textbar Broadcast signal\#1\textbar 
\newline
\textbar Communication Processor\textbar \quad \textbar directly\_performs\textbar \quad \textbar Manage communication\textbar 
\newline
\textbar Broadcast signal\#1\textbar \quad \textbar Data\textbar \quad \textbar Manage communication\textbar 
\newline
\textbar Broadcast signal\#1\textbar \quad \textbar Information from earth\textbar \quad \textbar Broadcast information from space\textbar 
\newline
\textbar Broadcast information from space\textbar \quad \textbar is\_performed\_by\textbar \quad \textbar Tracking \& Data Relay Satellite\textbar 
\newline
\textbar Communication module\textbar \quad \textbar Internal Protocol\textbar \quad \textbar Data Management module\textbar 
\newline
\textbar Support System Module\textbar \quad \textbar is\_contained\_by\textbar \quad \textbar HUBBLE Space Telescope\textbar 
\newline
[...]}}\\\hline
\rowcolor{lightpink}
\multicolumn{1}{p{14cm}}{\textbf{Baseline} \par
1. The system description includes two antennas (Antenna\#1) which are directly contained by the Communication module. 
\par
2. The Communication module directly contains the Communication Processor, which directly performs the action "Manage communication".
\par
3. The "Manage communication" action is performed by the Communication Processor, which implies that the Communication Processor is responsible for managing the communication process.
\par
4. The requirement states that the spacecraft shall ensure continuous communication with Earth by utilizing dual antennas to broadcast and receive signals, managed by an integrated Communication Processor.
        
\textbf{Based on the system description, we can conclude that the requirement is fulfilled. The system includes dual antennas (Antenna\#1) which are managed by the Communication Processor, ensuring continuous communication with Earth.}
Final Answer: Yes} \\
\hline
\rowcolor{lightgreen}
\multicolumn{1}{p{14cm}}{\textbf{Fine-Tuning} \par
1. The system description includes two antennas (Antenna\#1) which are directly contained by the Communication module. 
\par
2. The Communication module directly contains the Communication Processor, which directly performs the action "Manage communication".
\par
3. The "Manage communication" action is performed by the Communication Processor, which implies that the Communication Processor is responsible for managing the communication process.
\par
4. The requirement states that the spacecraft shall ensure continuous communication with Earth by utilizing dual antennas to broadcast and receive signals, managed by an integrated Communication Processor.
        
\textbf{Based on the system description, we can conclude that the requirement is fulfilled. The system includes dual antennas (Antenna\#1) which are managed by the Communication Processor, ensuring continuous communication with Earth. Final Answer: Yes}} \\
\hline
\rowcolor{lightgreen}
\multicolumn{1}{p{14cm}}{\textbf{Intervention} \par
1. The system description mentions two antennas: Antenna\#1 and Antenna\#2. However, only Antenna\#1 is directly contained by the Communication module. There is no mention of Antenna\#2.

\par
2. The requirement states that the spacecraft shall utilize dual antennas to broadcast and receive signals. Since only one antenna is mentioned in the system description, this requirement is not fulfilled.

\par
3. The requirement also mentions that the Communication Processor shall manage the communication. In the system description, the Communication Processor is directly contained by the Communication module and directly performs the action "Manage communication". 
        \textbf{However, it is not clear if the Communication Processor manages the communication for both antennas or only one. Final Answer: No}} \\

\hline
\end{tabular}
}
\end{table*}


\begin{table*}[!h]
{\scriptsize
\centering
\setlength{\tabcolsep}{3pt}
\caption{Example input and output for baseline, fine-tuned, and intervention models for requirement RQ-H-31. Incorrect outputs are in red; correct outputs are in green. Yellow output signifies correct answer but incorrect reasoning.}
\begin{tabular}{p{5.5cm}|p{5.5cm}|p{5.5cm}}
\hline
\multicolumn{3}{p{15cm}}{\textbf{Requirement: The Optical Telescope module shall align its instruments with the focal plane to provide wide field imagery in ultraviolet, visible, and infrared light.}} \\
\hline
\multicolumn{3}{p{15cm}}{\textit{Graph triples:
\newline
\textbar Optical Telescope module\textbar \quad \textbar directly\_contains\textbar \quad \textbar Focal Plane\textbar 
\newline
\textbar Focal Plane\textbar \quad \textbar directly\_performs\textbar \quad \textbar Align instruments with focal plane\textbar 
\newline
\textbar Align instruments with focal plane\textbar \quad \textbar Optical images\textbar \quad \textbar Provide wide field imagery in visible wavelengths\textbar 
\newline
\textbar Align instruments with focal plane\textbar \quad \textbar Optical images\textbar \quad \textbar Provide a wide field imagery in visible and infrared light\textbar 
\newline
\textbar Align instruments with focal plane\textbar \quad \textbar Optical images\textbar \quad \textbar Provide ultra violet spectrum\textbar 
\newline
\textbar Provide wide field imagery in visible wavelengths\textbar \quad \textbar is\_performed\_by\textbar \quad \textbar Wide Field Imagery sensor\#1\textbar 
\newline
\textbar Provide a wide field imagery in visible and infrared light\textbar \quad \textbar is\_performed\_by\textbar \quad \textbar Wide Field Imagery sensor\#2\textbar 
\newline
\textbar Provide ultra violet spectrum\textbar \quad \textbar is\_performed\_by\textbar \quad \textbar Ultra Violet sensor\textbar 
\newline
[...]}}\\\hline
\rowcolor{lightpink}
\multicolumn{1}{p{14cm}}{\textbf{Baseline} \par
1. The requirement is: "The Optical Telescope module shall align its instruments with the focal plane to provide wide field imagery in ultraviolet, visible, and infrared light."
\par
2. From the system description, we can see that the Optical Telescope module contains the Focal Plane, which directly performs the action "Align instruments with focal plane".
\par
3. The Focal Plane is also directly contained by the Optical Telescope module.
\par
4. The requirement mentions providing wide field imagery in ultraviolet, visible, and infrared light. 
\par
5. From the system description, we can see that the Focal Plane directly performs the actions "Provide wide field imagery in visible wavelengths" and "Provide a wide field imagery in visible and infrared light". \textbf{It also performs the actions "Provide ultra violet spectrum" and "Obtain high resolution spectra of resolved objetcs".}\par
\textbf{6. Therefore, the requirement is fulfilled because the Focal Plane, which is contained by the Optical Telescope module, performs the necessary actions to provide wide field imagery in ultraviolet, visible, and infrared light.}
\textbf{Final Answer: Yes}} \\
\hline
\rowcolor{lightyellow}
\multicolumn{1}{p{14cm}}{\textbf{Fine-Tuning} \par
1. The requirement is: "The Optical Telescope module shall align its instruments with the focal plane to provide wide field imagery in ultraviolet, visible, and infrared light."
\par
2. From the system description, we can see that the Optical Telescope module contains the Focal Plane, which directly performs the action "Align instruments with focal plane".
\par
3. The Focal Plane is contained by the Optical Telescope module and directly performs the action "Align instruments with focal plane".
\par
4. The Focal Plane also performs the actions "Provide wide field imagery in visible wavelengths", "Provide a wide field imagery in visible and infrared light", and "Provide ultra violet spectrum" through the sensors Wide Field Imagery sensor\#1, Wide Field Imagery sensor\#2, and Ultra Violet sensor respectively.
\par
5. The requirement mentions providing wide field imagery in ultraviolet, visible, and infrared light, which is partially fulfilled by the Focal Plane's actions.
\par
\textbf{6. However, the requirement does not explicitly mention the specific wavelengths or ranges of wavelengths, which are only mentioned in the actions performed by the Focal Plane.}
\textbf{Final Answer: No}
} \\
\hline
\rowcolor{lightgreen}
\multicolumn{1}{p{14cm}}{\textbf{Intervention} \par
1.	The requirement is: "The Optical Telescope module shall align its instruments with the focal plane to provide wide field imagery in ultraviolet, visible, and infrared light."
\par
2. From the system description, we can see that the Optical Telescope module contains the Focal Plane, which directly performs the action "Align instruments with focal plane".
\par
3. The Focal Plane is contained by the Optical Telescope module, and it directly performs the action "Align instruments with focal plane".
\par
4. The Focal Plane also contains the Wide Field Imagery sensor\#1, which provides wide field imagery in visible wavelengths, and Wide Field Imagery sensor\#2, which provides wide field imagery in visible and infrared light.
\par
\textbf{5. However, there is no mention of the Focal Plane providing wide field imagery in ultraviolet light. The Ultra Violet sensor is contained by the Instruments module, but it is not directly contained by the Focal Plane.}
\par
\textbf{6. Therefore, the requirement is not fully fulfilled because it asks for wide field imagery in ultraviolet, visible, and infrared light, but the system description only mentions wide field imagery in visible and infrared light.}
\textbf{Final Answer: No}
} \\

\hline
\end{tabular}
}
\end{table*}


\begin{table*}[!h]
{\scriptsize
\centering
\setlength{\tabcolsep}{3pt}
\caption{Example input and output for baseline, fine-tuned, and intervention models for requirement RQ-MB-3. Incorrect outputs are in red; correct outputs are in green.}
\begin{tabular}{p{5.5cm}|p{5.5cm}|p{5.5cm}}
\hline
\multicolumn{3}{p{15cm}}{\textbf{Requirement: The settlement shall maintain optimal temperature levels by utilizing the thermal control system to generate and reject heat as necessary, ensuring a stable environment for crew activities and habitation.}} \\
\hline
\multicolumn{3}{p{15cm}}{\textit{Graph triples:
\newline
\textbar Thermal Control\textbar \quad \textbar directly\_performs\textbar \quad \textbar Control Temperature\textbar 
\newline
\textbar Control Temperature\textbar \quad \textbar Generate Heat\textbar \quad \textbar Kitchen\textbar 
\newline
\textbar Control Temperature\textbar \quad \textbar Generate Heat\textbar \quad \textbar Execute Work\textbar 
\newline
\textbar Execute Work\textbar \quad \textbar is\_performed\_by\textbar \quad \textbar Crew\textbar 
\newline
\textbar Kitchen\textbar \quad \textbar is\_performed\_by\textbar \quad \textbar Habitation\textbar 
\newline
\textbar Control Temperature\textbar \quad \textbar is\_performed\_by\textbar \quad \textbar Thermal Control\textbar 
\newline
[...]}}\\\hline
\rowcolor{lightpink}
\multicolumn{1}{p{14cm}}{\textbf{Baseline} \par
1. The requirement states that the settlement shall maintain optimal temperature levels. 

\par
2. The system description shows that the Thermal Control system directly performs Control Temperature, which in turn generates heat and is contained by the Kitchen. 

\par
3. The Kitchen is performed by the Habitation, which is contained by the Logical System. 

\par
4. The Logical System contains the Thermal Control system, which means it has control over the Thermal Control system. 

\par
5. The requirement also states that the Thermal Control system shall generate and reject heat as necessary. 

\par
6. The system description shows that the Thermal Control system generates heat and is contained by the Kitchen, which is performed by the Habitation. 

\par
7. The Habitation is contained by the Logical System, which means it has control over the Habitation. 

\par
8. The Logical System contains the Thermal Control system, which means it has control over the Thermal Control system.

Based on the above analysis, the requirement is fulfilled. The Thermal Control system is capable of generating \textbf{and rejecting heat as necessary}, ensuring a stable environment for crew activities and habitation. 
\textbf{Final Answer: Yes}} \\
\hline
\rowcolor{lightpink}
\multicolumn{1}{p{14cm}}{\textbf{Fine-Tuning} \par
1. The requirement states that the settlement shall maintain optimal temperature levels. 

\par
2. The system description shows that the Thermal Control system directly performs Control Temperature, which generates heat and is contained by the Logical System.

\par
3. The Logical System contains the Thermal Control system, which means that the Thermal Control system is part of the Logical System.

\par
4. The system description also shows that the Thermal Control system directly performs Distribute Power, which is performed by the Power system.

\par
5. The requirement does not explicitly mention the distribution of power, but it does mention the maintenance of optimal temperature levels.

\textbf{Based on the given system description, it can be concluded that the Thermal Control system is responsible for generating and rejecting heat as necessary to maintain optimal temperature levels. The system description shows that the Thermal Control system directly performs Control Temperature, which generates heat and is contained by the Logical System.}

\textbf{Final Answer: Yes}
} \\
\hline
\rowcolor{lightgreen}
\multicolumn{1}{p{14cm}}{\textbf{Intervention} \par
1. The requirement states that the settlement shall maintain optimal temperature levels. 

\par
2. The system description shows that the Thermal Control system directly performs Control Temperature, which generates Heat and is contained by the Logical System.

\par
3. The Logical System contains the Thermal Control system, which means that the Thermal Control system is part of the Logical System.

\par
4. The requirement also states that the thermal control system shall generate and reject heat as necessary, ensuring a stable environment for crew activities and habitation.
\par
\textbf{5. The system description shows that the Thermal Control system generates Heat, but it does not show any mechanism for rejecting heat.}
\textbf{Final Answer: No}
} \\
\hline
\end{tabular}
}
\end{table*}


\FloatBarrier

\bibliographystyle{elsarticle-harv} 
\bibliography{bibliography}

\begin{thebibliography}{32}
\expandafter\ifx\csname natexlab\endcsname\relax\def\natexlab#1{#1}\fi
\providecommand{\url}[1]{\texttt{#1}}
\providecommand{\href}[2]{#2}
\providecommand{\path}[1]{#1}
\providecommand{\DOIprefix}{doi:}
\providecommand{\ArXivprefix}{arXiv:}
\providecommand{\URLprefix}{URL: }
\providecommand{\Pubmedprefix}{pmid:}
\providecommand{\doi}[1]{\href{http://dx.doi.org/#1}{\path{#1}}}
\providecommand{\Pubmed}[1]{\href{pmid:#1}{\path{#1}}}
\providecommand{\bibinfo}[2]{#2}
\ifx\xfnm\relax \def\xfnm[#1]{\unskip,\space#1}\fi
\bibitem[{Arditi et~al.(2024)Arditi, Obeso, Syed, Paleka, Panickssery, Gurnee and Nanda}]{arditi2024refusallanguagemodelsmediated}
\bibinfo{author}{Arditi, A.}, \bibinfo{author}{Obeso, O.}, \bibinfo{author}{Syed, A.}, \bibinfo{author}{Paleka, D.}, \bibinfo{author}{Panickssery, N.}, \bibinfo{author}{Gurnee, W.}, \bibinfo{author}{Nanda, N.}, \bibinfo{year}{2024}.
\newblock \bibinfo{title}{Refusal in language models is mediated by a single direction}.
\newblock \URLprefix \url{https://arxiv.org/abs/2406.11717}, \href{http://arxiv.org/abs/2406.11717}{{\tt arXiv:2406.11717}}.
\bibitem[{Berquand et~al.(2021)Berquand, Darm and Riccardi}]{berquandspacebert}
\bibinfo{author}{Berquand, A.}, \bibinfo{author}{Darm, P.}, \bibinfo{author}{Riccardi, A.}, \bibinfo{year}{2021}.
\newblock \bibinfo{title}{Spacetransformers: Language modeling for space systems}.
\newblock \bibinfo{journal}{IEEE Access} \bibinfo{volume}{9}, \bibinfo{pages}{133111--133122}.
\newblock \DOIprefix\doi{10.1109/ACCESS.2021.3115659}.
\bibitem[{Chang et~al.(2024)Chang, Wiens, Schnabel and Swaminathan}]{chang2024measuring}
\bibinfo{author}{Chang, T.}, \bibinfo{author}{Wiens, J.}, \bibinfo{author}{Schnabel, T.}, \bibinfo{author}{Swaminathan, A.}, \bibinfo{year}{2024}.
\newblock \bibinfo{title}{Measuring steerability in large language models}, in: \bibinfo{booktitle}{Neurips Safe Generative AI Workshop 2024}.
\newblock \URLprefix \url{https://openreview.net/forum?id=y2J5dAqcJW}.
\bibitem[{Cosler et~al.(2023)Cosler, Hahn, Mendoza, Schmitt and Trippel}]{nl2spec_cosler}
\bibinfo{author}{Cosler, M.}, \bibinfo{author}{Hahn, C.}, \bibinfo{author}{Mendoza, D.}, \bibinfo{author}{Schmitt, F.}, \bibinfo{author}{Trippel, C.}, \bibinfo{year}{2023}.
\newblock \bibinfo{title}{nl2spec: Interactively translating unstructured natural language to temporal logics with large language models}, in: \bibinfo{editor}{Enea, C.}, \bibinfo{editor}{Lal, A.} (Eds.), \bibinfo{booktitle}{Computer Aided Verification}, \bibinfo{publisher}{Springer Nature Switzerland}, \bibinfo{address}{Cham}. pp. \bibinfo{pages}{383--396}.
\bibitem[{Darm et~al.(2023)Darm, Marchetti, Garcia, Redondo, Riccardi and Fern{\'a}ndez}]{darm_semantic_similarity}
\bibinfo{author}{Darm, P.}, \bibinfo{author}{Marchetti, F.}, \bibinfo{author}{Garcia, G.}, \bibinfo{author}{Redondo, P.}, \bibinfo{author}{Riccardi, A.}, \bibinfo{author}{Fern{\'a}ndez, A.}, \bibinfo{year}{2023}.
\newblock \bibinfo{title}{Leveraging language models semantic similarity capabilities to facilitate information reuse in system engineering}.
\newblock \URLprefix \url{https://www.iac2023.org/}. \bibinfo{note}{74th International Astronautical Congress, IAC 2023 ; Conference date: 02-10-2023 Through 06-10-2023}.
\bibitem[{Darm and Riccardi(2025)}]{darm2025letaiconspiracybegin}
\bibinfo{author}{Darm, P.}, \bibinfo{author}{Riccardi, A.}, \bibinfo{year}{2025}.
\newblock \bibinfo{title}{"let the ai conspiracy begin..." language model coordination is just one inference-intervention away}.
\newblock \URLprefix \url{https://arxiv.org/abs/2502.05945}, \href{http://arxiv.org/abs/2502.05945}{{\tt arXiv:2502.05945}}.
\bibitem[{Ethayarajh et~al.(2024)Ethayarajh, Xu, Muennighoff, Jurafsky and Kiela}]{ethayarajh_kahneman}
\bibinfo{author}{Ethayarajh, K.}, \bibinfo{author}{Xu, W.}, \bibinfo{author}{Muennighoff, N.}, \bibinfo{author}{Jurafsky, D.}, \bibinfo{author}{Kiela, D.}, \bibinfo{year}{2024}.
\newblock \bibinfo{title}{Model alignment as prospect theoretic optimization}, in: \bibinfo{booktitle}{Proceedings of the 41st International Conference on Machine Learning}, \bibinfo{publisher}{JMLR.org}.
\bibitem[{Franch et~al.(2023)Franch, Palomares, Quer, Chatzipetrou and Gorschek}]{Franch2023}
\bibinfo{author}{Franch, X.}, \bibinfo{author}{Palomares, C.}, \bibinfo{author}{Quer, C.}, \bibinfo{author}{Chatzipetrou, P.}, \bibinfo{author}{Gorschek, T.}, \bibinfo{year}{2023}.
\newblock \bibinfo{title}{The state-of-practice in requirements specification: an extended interview study at 12 companies}.
\newblock \bibinfo{journal}{Requirements Engineering} \bibinfo{volume}{28}, \bibinfo{pages}{377--409}.
\newblock \URLprefix \url{https://doi.org/10.1007/s00766-023-00399-7}, \DOIprefix\doi{10.1007/s00766-023-00399-7}.
\bibitem[{Hause et~al.(2006)}]{hause2006sysml}
\bibinfo{author}{Hause, M.}, et~al., \bibinfo{year}{2006}.
\newblock \bibinfo{title}{The sysml modelling language}, in: \bibinfo{booktitle}{Fifteenth European Systems Engineering Conference}, pp. \bibinfo{pages}{1--12}.
\bibitem[{Hu et~al.(2021)Hu, Shen, Wallis, Allen-Zhu, Li, Wang, Wang and Chen}]{hu2021loralowrankadaptationlarge}
\bibinfo{author}{Hu, E.J.}, \bibinfo{author}{Shen, Y.}, \bibinfo{author}{Wallis, P.}, \bibinfo{author}{Allen-Zhu, Z.}, \bibinfo{author}{Li, Y.}, \bibinfo{author}{Wang, S.}, \bibinfo{author}{Wang, L.}, \bibinfo{author}{Chen, W.}, \bibinfo{year}{2021}.
\newblock \bibinfo{title}{Lora: Low-rank adaptation of large language models}.
\newblock \URLprefix \url{https://arxiv.org/abs/2106.09685}, \href{http://arxiv.org/abs/2106.09685}{{\tt arXiv:2106.09685}}.
\bibitem[{INCOSE(2022)}]{INCOSE2022}
\bibinfo{author}{INCOSE}, \bibinfo{year}{2022}.
\newblock \bibinfo{title}{INCOSE Systems Engineering Vision 2035. Engineering solutions for a better world}.
\bibitem[{{INCOSE Technical Operations}(2007)}]{INCOSE2007}
\bibinfo{author}{{INCOSE Technical Operations}}, \bibinfo{year}{2007}.
\newblock \bibinfo{title}{Systems Engineering Vision 2020, version 2.03}.
\newblock \bibinfo{type}{Technical Report} \bibinfo{number}{INCOSE-TP-2004-004-02}. International Council on Systems Engineering (INCOSE). \bibinfo{address}{Seattle, WA}.
\bibitem[{Jorgensen et~al.(2023)Jorgensen, Cope, Schoots and Shanahan}]{jorgensen2023improvingactivationsteeringlanguage}
\bibinfo{author}{Jorgensen, O.}, \bibinfo{author}{Cope, D.}, \bibinfo{author}{Schoots, N.}, \bibinfo{author}{Shanahan, M.}, \bibinfo{year}{2023}.
\newblock \bibinfo{title}{Improving activation steering in language models with mean-centring}.
\newblock \URLprefix \url{https://arxiv.org/abs/2312.03813}, \href{http://arxiv.org/abs/2312.03813}{{\tt arXiv:2312.03813}}.
\bibitem[{Li et~al.(2023)Li, Patel, Vi{\'e}gas, Pfister and Wattenberg}]{li2023inferencetime}
\bibinfo{author}{Li, K.}, \bibinfo{author}{Patel, O.}, \bibinfo{author}{Vi{\'e}gas, F.}, \bibinfo{author}{Pfister, H.}, \bibinfo{author}{Wattenberg, M.}, \bibinfo{year}{2023}.
\newblock \bibinfo{title}{Inference-time intervention: Eliciting truthful answers from a language model}, in: \bibinfo{booktitle}{Thirty-seventh Conference on Neural Information Processing Systems}.
\newblock \URLprefix \url{https://openreview.net/forum?id=aLLuYpn83y}.
\bibitem[{Madni and Purohit(2019)}]{Madni2019}
\bibinfo{author}{Madni, A.M.}, \bibinfo{author}{Purohit, S.}, \bibinfo{year}{2019}.
\newblock \bibinfo{title}{Economic analysis of model-based systems engineering}.
\newblock \bibinfo{journal}{Systems} \bibinfo{volume}{7}.
\newblock \DOIprefix\doi{10.3390/systems7010012}.
\bibitem[{Marks and Tegmark(2024)}]{marks2024geometrytruthemergentlinear}
\bibinfo{author}{Marks, S.}, \bibinfo{author}{Tegmark, M.}, \bibinfo{year}{2024}.
\newblock \bibinfo{title}{The geometry of truth: Emergent linear structure in large language model representations of true/false datasets}.
\newblock \URLprefix \url{https://arxiv.org/abs/2310.06824}, \href{http://arxiv.org/abs/2310.06824}{{\tt arXiv:2310.06824}}.
\bibitem[{Miehling et~al.(2025)Miehling, Desmond, Ramamurthy, Daly, Dognin, Rios, Bouneffouf and Liu}]{miehling2025evaluatingpromptsteerabilitylarge}
\bibinfo{author}{Miehling, E.}, \bibinfo{author}{Desmond, M.}, \bibinfo{author}{Ramamurthy, K.N.}, \bibinfo{author}{Daly, E.M.}, \bibinfo{author}{Dognin, P.}, \bibinfo{author}{Rios, J.}, \bibinfo{author}{Bouneffouf, D.}, \bibinfo{author}{Liu, M.}, \bibinfo{year}{2025}.
\newblock \bibinfo{title}{Evaluating the prompt steerability of large language models}.
\newblock \URLprefix \url{https://arxiv.org/abs/2411.12405}, \href{http://arxiv.org/abs/2411.12405}{{\tt arXiv:2411.12405}}.
\bibitem[{Norheim et~al.(2024)Norheim, Rebentisch, Xiao, Draeger, Kerbrat and de~Weck}]{Norheim_2024}
\bibinfo{author}{Norheim, J.J.}, \bibinfo{author}{Rebentisch, E.}, \bibinfo{author}{Xiao, D.}, \bibinfo{author}{Draeger, L.}, \bibinfo{author}{Kerbrat, A.}, \bibinfo{author}{de~Weck, O.L.}, \bibinfo{year}{2024}.
\newblock \bibinfo{title}{Challenges in applying large language models to requirements engineering tasks}.
\newblock \bibinfo{journal}{Design Science} \bibinfo{volume}{10}, \bibinfo{pages}{e16}.
\newblock \DOIprefix\doi{10.1017/dsj.2024.8}.
\bibitem[{Panickssery et~al.(2024)Panickssery, Gabrieli, Schulz, Tong, Hubinger and Turner}]{panickssery2024steeringllama2contrastive}
\bibinfo{author}{Panickssery, N.}, \bibinfo{author}{Gabrieli, N.}, \bibinfo{author}{Schulz, J.}, \bibinfo{author}{Tong, M.}, \bibinfo{author}{Hubinger, E.}, \bibinfo{author}{Turner, A.M.}, \bibinfo{year}{2024}.
\newblock \bibinfo{title}{Steering llama 2 via contrastive activation addition}.
\newblock \URLprefix \url{https://arxiv.org/abs/2312.06681}, \href{http://arxiv.org/abs/2312.06681}{{\tt arXiv:2312.06681}}.
\bibitem[{Praehofer and Kerschbaummayr(1999)}]{PRAEHOFER1999717}
\bibinfo{author}{Praehofer, H.}, \bibinfo{author}{Kerschbaummayr, J.}, \bibinfo{year}{1999}.
\newblock \bibinfo{title}{Case-based reasoning techniques to support reusability in a requirement engineering and system design tool}.
\newblock \bibinfo{journal}{Engineering Applications of Artificial Intelligence} \bibinfo{volume}{12}, \bibinfo{pages}{717--731}.
\newblock \URLprefix \url{https://www.sciencedirect.com/science/article/pii/S0952197699000433}, \DOIprefix\doi{https://doi.org/10.1016/S0952-1976(99)00043-3}.
\bibitem[{Qiu et~al.(2024)Qiu, Zhao, Ziser, Korhonen, Ponti and Cohen}]{qiu2024spectral}
\bibinfo{author}{Qiu, Y.}, \bibinfo{author}{Zhao, Z.}, \bibinfo{author}{Ziser, Y.}, \bibinfo{author}{Korhonen, A.}, \bibinfo{author}{Ponti, E.M.}, \bibinfo{author}{Cohen, S.B.}, \bibinfo{year}{2024}.
\newblock \bibinfo{title}{Spectral editing of activations for large language model alignment}.
\newblock \href{http://arxiv.org/abs/2405.09719}{{\tt arXiv:2405.09719}}.
\bibitem[{Rafailov et~al.(2024)Rafailov, Sharma, Mitchell, Ermon, Manning and Finn}]{rafailov2024directpreferenceoptimizationlanguage}
\bibinfo{author}{Rafailov, R.}, \bibinfo{author}{Sharma, A.}, \bibinfo{author}{Mitchell, E.}, \bibinfo{author}{Ermon, S.}, \bibinfo{author}{Manning, C.D.}, \bibinfo{author}{Finn, C.}, \bibinfo{year}{2024}.
\newblock \bibinfo{title}{Direct preference optimization: Your language model is secretly a reward model}.
\newblock \URLprefix \url{https://arxiv.org/abs/2305.18290}, \href{http://arxiv.org/abs/2305.18290}{{\tt arXiv:2305.18290}}.
\bibitem[{Rogers and Mitchell(2021)}]{Rogers2021}
\bibinfo{author}{Rogers, E.B.}, \bibinfo{author}{Mitchell, S.W.}, \bibinfo{year}{2021}.
\newblock \bibinfo{title}{Mbse delivers significant return on investment in evolutionary development of complex sos}.
\newblock \bibinfo{journal}{Systems Engineering} \bibinfo{volume}{24}.
\newblock \DOIprefix\doi{10.1002/sys.21592}.
\bibitem[{Tikayat~Ray et~al.(2023)Tikayat~Ray, Cole, Pinon~Fischer, White and Mavris}]{ray_aeroBERT}
\bibinfo{author}{Tikayat~Ray, A.}, \bibinfo{author}{Cole, B.F.}, \bibinfo{author}{Pinon~Fischer, O.J.}, \bibinfo{author}{White, R.T.}, \bibinfo{author}{Mavris, D.N.}, \bibinfo{year}{2023}.
\newblock \bibinfo{title}{aerobert-classifier: Classification of aerospace requirements using bert}.
\newblock \bibinfo{journal}{Aerospace} \bibinfo{volume}{10}.
\newblock \URLprefix \url{https://www.mdpi.com/2226-4310/10/3/279}, \DOIprefix\doi{10.3390/aerospace10030279}.
\bibitem[{Topcu et~al.(2025)Topcu, Husain, Ofsa and Wach}]{topcu2025trustperilmixedmethods}
\bibinfo{author}{Topcu, T.G.}, \bibinfo{author}{Husain, M.}, \bibinfo{author}{Ofsa, M.}, \bibinfo{author}{Wach, P.}, \bibinfo{year}{2025}.
\newblock \bibinfo{title}{Trust at your own peril: A mixed methods exploration of the ability of large language models to generate expert-like systems engineering artifacts and a characterization of failure modes}.
\newblock \URLprefix \url{https://arxiv.org/abs/2502.09690}, \href{http://arxiv.org/abs/2502.09690}{{\tt arXiv:2502.09690}}.
\bibitem[{Vaswani et~al.(2017)Vaswani, Shazeer, Parmar, Uszkoreit, Jones, Gomez, Kaiser and Polosukhin}]{vaswani2017attention}
\bibinfo{author}{Vaswani, A.}, \bibinfo{author}{Shazeer, N.}, \bibinfo{author}{Parmar, N.}, \bibinfo{author}{Uszkoreit, J.}, \bibinfo{author}{Jones, L.}, \bibinfo{author}{Gomez, A.N.}, \bibinfo{author}{Kaiser, {\L}.}, \bibinfo{author}{Polosukhin, I.}, \bibinfo{year}{2017}.
\newblock \bibinfo{title}{Attention is all you need}.
\newblock \bibinfo{journal}{Advances in neural information processing systems} \bibinfo{volume}{30}.
\bibitem[{Voirin(2017)}]{voirin2017model}
\bibinfo{author}{Voirin, J.L.}, \bibinfo{year}{2017}.
\newblock \bibinfo{title}{Model-based System and Architecture Engineering with the Arcadia Method}.
\newblock \bibinfo{edition}{1} ed., \bibinfo{publisher}{ISTE Press - Elsevier}, \bibinfo{address}{London}.
\newblock \bibinfo{note}{Hardback ISBN: 978-1-78548-169-7, eBook ISBN: 9780081017944}.
\bibitem[{Wach and Salado(2022)}]{salado_syml_extension}
\bibinfo{author}{Wach, P.}, \bibinfo{author}{Salado, A.}, \bibinfo{year}{2022}.
\newblock \bibinfo{title}{The need for semantic extension of sysml to model the problem space}, in: \bibinfo{editor}{Madni, A.M.}, \bibinfo{editor}{Boehm, B.}, \bibinfo{editor}{Erwin, D.}, \bibinfo{editor}{Moghaddam, M.}, \bibinfo{editor}{Sievers, M.}, \bibinfo{editor}{Wheaton, M.} (Eds.), \bibinfo{booktitle}{Recent Trends and Advances in Model Based Systems Engineering}, \bibinfo{publisher}{Springer International Publishing}, \bibinfo{address}{Cham}. pp. \bibinfo{pages}{279--289}.
\bibitem[{Wang et~al.(2019)Wang, Zhong, Zhao and Ren}]{wang_model_check_sysml}
\bibinfo{author}{Wang, H.}, \bibinfo{author}{Zhong, D.}, \bibinfo{author}{Zhao, T.}, \bibinfo{author}{Ren, F.}, \bibinfo{year}{2019}.
\newblock \bibinfo{title}{Integrating model checking with sysml in complex system safety analysis}.
\newblock \bibinfo{journal}{IEEE Access} \bibinfo{volume}{7}, \bibinfo{pages}{16561--16571}.
\newblock \DOIprefix\doi{10.1109/ACCESS.2019.2892745}.
\bibitem[{Wang et~al.(2023)Wang, Wei, Schuurmans, Le, Chi, Narang, Chowdhery and Zhou}]{wang2023selfconsistencyimproveschainthought}
\bibinfo{author}{Wang, X.}, \bibinfo{author}{Wei, J.}, \bibinfo{author}{Schuurmans, D.}, \bibinfo{author}{Le, Q.}, \bibinfo{author}{Chi, E.}, \bibinfo{author}{Narang, S.}, \bibinfo{author}{Chowdhery, A.}, \bibinfo{author}{Zhou, D.}, \bibinfo{year}{2023}.
\newblock \bibinfo{title}{Self-consistency improves chain of thought reasoning in language models}.
\newblock \URLprefix \url{https://arxiv.org/abs/2203.11171}, \href{http://arxiv.org/abs/2203.11171}{{\tt arXiv:2203.11171}}.
\bibitem[{Wei et~al.(2023)Wei, Wang, Schuurmans, Bosma, Ichter, Xia, Chi, Le and Zhou}]{wei2023chainofthoughtpromptingelicitsreasoning}
\bibinfo{author}{Wei, J.}, \bibinfo{author}{Wang, X.}, \bibinfo{author}{Schuurmans, D.}, \bibinfo{author}{Bosma, M.}, \bibinfo{author}{Ichter, B.}, \bibinfo{author}{Xia, F.}, \bibinfo{author}{Chi, E.}, \bibinfo{author}{Le, Q.}, \bibinfo{author}{Zhou, D.}, \bibinfo{year}{2023}.
\newblock \bibinfo{title}{Chain-of-thought prompting elicits reasoning in large language models}.
\newblock \URLprefix \url{https://arxiv.org/abs/2201.11903}, \href{http://arxiv.org/abs/2201.11903}{{\tt arXiv:2201.11903}}.
\bibitem[{Xu et~al.(2024)Xu, Huang, Wang, Wu, Yao and Xie}]{xu2024uncovering}
\bibinfo{author}{Xu, Z.}, \bibinfo{author}{Huang, R.}, \bibinfo{author}{Wang, X.}, \bibinfo{author}{Wu, F.}, \bibinfo{author}{Yao, J.}, \bibinfo{author}{Xie, X.}, \bibinfo{year}{2024}.
\newblock \bibinfo{title}{Uncovering safety risks in open-source llms through concept activation vector}.
\newblock \bibinfo{journal}{arXiv preprint arXiv:2404.12038} .

\end{thebibliography}

\end{document}